\documentclass[11pt]{elsarticle}
 \journal{******}
\usepackage{appendix}
\usepackage{bbding}
\usepackage{amsfonts}
\usepackage{color}
\usepackage{mathrsfs}
\usepackage{rotating}
\usepackage{graphicx}
\usepackage{epstopdf}
\usepackage{subfigure}
\usepackage{caption}
\usepackage{float}
\usepackage{algorithm}
\usepackage[noend]{algpseudocode}
\usepackage{bm}
\usepackage{booktabs}
\usepackage{footnote}
\usepackage[colorlinks,
            linkcolor=blue,
            anchorcolor=red ,
            citecolor=blue
            ]{hyperref}

\usepackage{amsmath,amssymb, amsbsy, amsthm, array, amsfonts,
amssymb,cases,graphicx} \makeatletter
\usepackage{setspace}

\setcitestyle{authoryear,round}
\newcommand{\Extend}[5]{\ext@arrow 0099{\arrowfill@#1#2#3}{#4}{#5}}
\makeatother
\def\thebibliography#1{\leftline{\bf\normalsize References}\list
{[\arabic{enumi}]}{\settowidth\labelwidth{[#1]}\leftmargin\labelwidth
\advance\leftmargin\labelsep
\usecounter{enumi}}
\def\newblock{\hskip .11em plus .33em minus .07em}
\sloppy\clubpenalty4000\widowpenalty4000 \sfcode`\.=1000\relax}
\begin {document}

\newtheorem{theorem}{Theorem}[section]

\newtheorem{assumption}{Assumption}[section]

\newtheorem{example}{Example}[section]

\newtheorem{lemma}{Lemma}[section]

\newtheorem{remark}{Remark}[section]

\setlength\arraycolsep{1pt}
\def\S{\mathcal{S}}
\def\A{\mathcal{A}}
\def\E{\mathrm{E}}
\def\P{\mathrm{P}}

\def\qed{\hfill$\square$\smallskip}
\def\beqlb{\begin{eqnarray}}\def\eeqlb{\end{eqnarray}}
\def\beqnn{\begin{eqnarray*}}\def\eeqnn{\end{eqnarray*}}
\DeclareRobustCommand{\annu}[1]{_{%
\def\arraystretch{0}%
\setlength\arraycolsep{1pt}
\setlength\arrayrulewidth{.2pt}
}} \pagestyle{plain}
\begin{frontmatter}
\title{\bf Minimax Weight Learning for Absorbing MDPs\footnote{This research is supported by National Key R\&D Program of China (Nos. 2021YFA1000100 and 2021YFA1000101) and Natural Science Foundation of China (No. 71771089)}}
\author{Fengying Li, Yuqiang Li and Xianyi Wu\footnote{Corresponding authors:
Yuqiang Li at yqli@stat.ecnu.edu.cn and Xianyi Wu at xywu@stat.ecnu.edu.cn.}}

\address{School of Statistics, KLATASDS-MOE, East China Normal University, \\ Shanghai 200062, PR China}
\begin{abstract}
 {Reinforcement learning policy evaluation problems are often modeled as finite or discounted/averaged infinite-horizon Markov Decision Processes (MDPs). In this paper, we study undiscounted off-policy evaluation for absorbing MDPs. Given the dataset consisting of i.i.d episodes under a given truncation level, we propose an algorithm (referred to as MWLA in the text) to directly estimate the expected return via the
importance ratio of the state-action occupancy measure. 
The Mean Square Error (MSE) bound of the MWLA method is provided and the dependence of statistical errors on the data size and the truncation level are analyzed. The performance of the algorithm is illustrated by means of computational experiments under an episodic taxi environment}
\vskip3mm

\end{abstract}
\end{frontmatter}

\section{Introduction}

Off-policy evaluation (OPE) in reinforcement learning refers to the estimation of the expected returns of target policies using data collected by possibly different behavior policies. It is particularly important in situations where implementing new strategies is expensive, risky, or even dangerous, such as medicine (\citealp{murphy2001marginal}), education (\citealp{mandel2014offline}), economics (\citealp{hirano2003efficient}), recommender systems (\citealp{li2011unbiased}), and more. Currently available OPE procedures are mostly based on direct importance sampling (IS) techniques, which suffer from high variances that exponentially increase over the time horizon, known as "the curse of horizon" (\citealp{jiang2016doubly}; \citealp{li2015toward}).

A promising idea recently proposed uses marginalized importance sampling (MIS) to alleviate the curse of horizon, but raises a new problem of how to estimate the maginalized importance ratios. For instance, in the case of an infinite-horizon discounted Markov Decision Process (MDP), \cite{liu2018breaking} compute the importance weights based on the state distribution by solving a minimax optimization problem, and propose a method to estimate the expected return. Moreover, \cite{uehara2020minimax} propose a Minimax Weight Learning (MWL) algorithm that directly estimates the ratio of the state-action distribution without relying on the specification of the behavior policy.

In many real practices, such as robotic tasks, the environments will terminate at certain random times when they evolve into certain states. In such situations, it is no longer appropriate to model the environments using only finite-time or infinite-time MDPs. Instead, MDPs with absorbing states are suitable, where the absorption reflects the termination of the processes. Furthermore, from a theoretical perspective, absorbing MDPs extend the framework of infinite-horizon discounted MDP processes \citep{altman1999constrained} but the reverse does not hold (see Section \ref{rem2.1} for more details).

The theory of absorbing MDPs has been extensively studied and is well understood. For example, the knowledge of the times required to reach the absorbing states, depending on both the state and the actions, can be found in \cite{chatterjee2008stochastic} and \cite{iida1996markov} and the minimization of the expected undiscounted cost until the state enters the absorbing set in various applications (such as pursuit problems, transient programming, and first pass problems) are investigated by \cite{eaton1962optimal}, \cite{derman1970finite}, \cite{kushner1971introduction}), among others. Other research efforts include the stochastic shortest path problem (\citealp{bertsekas1991analysis}), the control-to-exit time problem (\citealp{kesten1975controlled}; \citealp{borkar1988convex}), among a vast number of others.

{In the context of learning when the underlying distributions are unknown, however, while many benchmark environments are indeed episodic and have random horizons, such as board games (a game terminates once the winner is determined), trips through a maze, and dialog systems (a session terminates when the conversation is concluded) (\citealp{jiang2017theory}), there are only limited efforts specifically contributed to absorbing MDPs. }
In this paper, we propose an MWL algorithm for offline RL involving absorbing MDPs, referred to as MWLA hereafter.  Our proposed approach tackles two key challenges.

The first challenge pertains to the Data structure. While an infinite horizon MDP can be treated as an ergodic Markov chain under suitable assumptions, the same assumption is not always valid for absorbing MDPs due to their indefinite horizons and varying episode lengths, of which some can be quite short. Therefore, assuming that the collected data consists of i.i.d. tuples $(s_t, a_t, r_t, s_{t+1})$, which is crucial for the MWL method, is not always natural. Instead, we propose working with data that consists of trajectories, where each data point represents a single trajectory.

The second challenge arises from the random episodic length and the expected undiscounted total rewards. In absorbing MDPs, the length of an episode is indefinite, and it may not be always practical to observe extremely long episodes fully, due to various reasons, such as their length or cost. To address this issue, a simple but practical choice is to truncate long episodes with a level $H$. If the expected total rewards are discounted with $\gamma<1$, it is easy to see that we can control the errors resulting from such truncations by sufficiently small $\gamma^H$. However, with the expected undiscounted total rewards, how to quantify the errors is unclear up to now for absorbing MDPs.

Our proposed MWLA algorithm deals with truncated episode data, and we provide a theoretical analysis of the errors resulting from episode truncation. As a result, it aids us in gaining a better understanding of the effects of episode truncation and identifying an appropriate truncation level under which the errors caused by truncation can be deemed acceptable.



Specifically, in this paper, we derive an estimate of the expected undiscounted return of an absorbing MDP and establish an upper bound for the MSE of the MWLA algorithm. The bound consists of three parts: errors due to statistics, approximation and optimization, respectively. The statistical error depends on both the truncation level and data size. Moreover, we present a uniform bound on MSE by means of an optimization with respect to the truncation level when the truncation level is relatively large. We also demonstrate the effectiveness of our algorithm through numerical experiments in the episodic taxi environment.

The remainder of this paper is organized as follows:
Section \ref{section2} introduces the model formulation and specifies some basic settings. The MWLA algorithm and its theoretical guarantees without knowledge of behavior policies are presented in Section \ref{section3}.  Additionally, we discuss a parallel version of MWLA, referred to MSWLA, for absorbing MDP with a known behavior policy in Remark \ref{MSWLA}.
In Section \ref{section4}, under the assumption that the data consists of i.i.d. episodes, MSE bound for the MWLA method is provided in Theorems \ref{Theorem 3} and \ref{Corollary 4}. Specifically, when the function classes are VC classes, compared with Theorem 9 in \cite{uehara2020minimax}, it is found that our statistical error is related to the truncation length $H$.
The related work is discussed in Section \ref{section5}, providing more details to clarify their connection to and differences between the current work.
In Section \ref{section6}, a computer experiment is reported under the episodic taxi environment, compared with on-policy, naive-average, and MSWLA methods, where estimates of returns and their MSEs are given under different episode numbers and truncation lengths.
All theoretical proofs and the pseudo-code of the algorithm are deferred to the Appendix.

\section{Basic setting}\label{section2}

An MDP is a controllable rewarded Markov process, represented by a tuple $ M=(\mathcal{S}, \mathcal{A}, \mathcal{R}, P, \mu)$ of a state space $\mathcal{S}$, an action space $\mathcal{A}$, a reward distribution $\mathcal R$ mapping a state-action pair $(s,a)$ to a distribution $\mathcal{R}(s,a)$ over the set $\mathbb{R}$ of real numbers with an expectation value $R(s, a)$, a transfer probability function $P:(s,a,s')\in\mathcal{S}\times \mathcal{A} \times \mathcal{S}\rightarrow P(s'|s,a)\in[0,1]$ and an initial state distribution $\mu$.

In this paper, a policy $\pi:=\pi(a|s)$  refers to a time-homogeneous mapping from $\mathcal{S}$ to the family of all distributions over $\mathcal{A}$, executed as follows. Starting with an initial state $s_{0}\sim \mu $, at any integer time $t\geq 0$, an action $a_{t}\sim\pi(\cdot|s_{t})$ is sampled, a scalar reward $r_{t}\sim\mathcal{R}(s_{t},a_{t})$ is collected, and a next state $s_{t+1}\sim P(\cdot | s_{t},a_{t})$ is then assigned by the environment. The space $\mathcal{S}\times\mathcal{A}$ is assumed enumerable and $R$ is bounded.  The probability distribution generated by $M$ under a policy $\pi$ and an initial distribution $\mu$  is denoted by $\P_{\mu, \pi}$, and $\E_{\mu,\pi}$ is used for its expectation operation. When the initial state $s_{0}=i$, the probability distribution and expectation are indicated by $\P_{i,\pi}$ and $\E_{i,\pi}$ respectively, so that $\P_{\mu,\pi}=\sum_{i\in \mathcal{S}}\mu(i)\P_{i,\pi}$ and $\E_{\mu,\pi}=\sum_{i\in\mathcal{S}}\mu(i)\E_{i,\pi}.$
The notation $\P_{(s,a),\pi}$ and $\E_{(s,a),\pi}$ are also used to indicate the probability and expectation generated by $M$ starting from the state-action couple $(s, a)$ and subsequently the following policy $\pi$.


An absorbing state, represented by $\xi\in \mathcal{S}$, is a state such that $r(\xi, a)=0$ and $P(\xi|\xi,a)=1$ for all $a\in \mathcal{A}$. It is assumed that there is only one unique absorbing state.  For a trajectory, denote by $T\doteq\min\{n\geq 1, s_n=\xi\}$ the terminal time, where and throughout the paper $\doteq$ signifies ``defined as''. An MDP is absorbing if  $\P_{i,\pi}(T<\infty)=1$ for all states $i$ and all policies $\pi$. Denote by $\mathcal{S}_0=\mathcal{S}\setminus\{\xi\}$ the non-absorbing states. We make the following assumption on $T$.

\begin{assumption}\label{Assumption 1} $\sup_{(s,a)\in\S_0\times\A,\pi}\E_{(s, a),\pi}(T)<\infty$.
\end{assumption}

Write $\Xi=\{ q:\S_0\times\A\mapsto \mathbb{R}\}$ for the collection of all functions mapping a couple \((s,a)\) to a real number. Introduce a functional $\Phi_\pi(q)$ on $\Xi$  by
 $$\Phi_\pi(q)\doteq\E_{\mu,\pi}[\sum_{t=0}^{T-1} q(s_t, a_t)]$$ and for any functions $q:\S_0\times\A\mapsto \mathbb{R}\), let $q(s,\pi)\dot=\sum_{a\in\A}\pi(a|s)q(s, a)$.
  Note that $\Phi_\pi$ implicitly depends on the initial distribution $\mu$.

The expected return under a policy $\pi$ is given by
\begin{equation}\label{objective function}
R_\pi\doteq\E_{\mu,\pi}\left[\sum\limits_{t=0}^{T-1}r_{t}\right]=\E_{\mu,\pi}\sum\limits_{t=0}^{\infty}\left[ R(s_t,a_t)\right]=\Phi_\pi(R),
\end{equation}
which depends only on the transition probability and the mean reward function rather than the more detailed reward distribution functions.

For any $(s, a)\in \S_0\times\A$, taking the indicator function ${\bm 1}_{(s,a)}$ as a particular mean reward function, i.e., collecting a unit reward at $(s,a)$ and zero otherwise, gives rise to the occupancy measure
\begin{equation}\label{a1}
d_{\pi}(s, a)
\doteq\mathrm{E}_{\mu,\pi}\big[\sum_{t=0}^{T-1}{\bm 1}_{(s, a)}(s_{t}, a_t)\big]=\sum_{t=0}^{\infty}\P_{\mu,\pi}(s_{t}=s, a_{t}=a)=\Phi_{\pi}({\bf 1}_{(s, a)}),
\end{equation}
where  $d_{\pi}(s, a)<\infty$ by Assumptions \ref{Assumption 1}. Conceptually, $d_{\pi}$ can also be retrieved from Chapmann-Kolmogorov equations
\beqlb\label{U1}
d_{\pi}(s, a)&=&
\mu(s)\pi(a|s)+\sum_{(s',a')\in \S_0\times\A}P(s|s',a')\pi(s|a)d_\pi(s', a'), \hbox{ for all }(s, a)\in\S_0\times\A.
\eeqlb
Conversely, with the measure $d_\pi$, we can retrieve the expected return
\beqnn
    R_\pi=\sum_{(s,a)\in S_0\times\A} R(s,a)d_{\pi}(s,a),
\eeqnn
as an integration of the mean reward function with respect to the occupancy measure.
In addition,
a simple recursive argument shows that
\beqnn
\Phi_\pi(q)&=&\sum_{(s,a)\in S_0\times\A} q(s,a)d_{\pi}(s,a)\notag\\
&=&\sum_{s\in\S_0}\mu(s)q(s,{\pi})+\sum_{(s,a,s')\in\S_0\times\A\times\S_0}P(s'|s,a)q(s',{\pi})d_{{\pi}}(s,a).
\eeqnn
A direct result of this equality is a family of equations
\beqlb\label{R_q_1}
\sum_{(s,a)\in S_0\times\A}\left[ q(s,a)-\sum_{s'\in\mathcal{S}_0}P(s'|s,a)q(s',{\pi})\right]d_{\pi}(s,a)
=\sum_{s'\in\S_0}\mu(s')q(s',{\pi}), q\in\Xi.
\eeqlb
Denote  by $d_{\pi}(s,a,s')=d_{\pi}(s,a)P(s'|s,a)$. Then the system of equations \eqref{R_q_1} can equivalently be rewritten as
\beqlb\label{f4}
\sum_{(s,a,s')\in\S_0\times\A\times\S_0}q(s',{\pi})d_{{\pi}}(s,a,s')-\sum_{(s,a)\in\S_0\times\A}q(s,a)d_{{\pi}}(s,a)+\mathrm{E}_{s\sim \mu}\big[q(s,{\pi})\big]=0, q\in\Xi.
\eeqlb

{ \begin{remark}\label{rem2.1}MDP models with absorbing states to maximize expected return \eqref{objective function} fundamentally differ from the usually analyzed  discounted MDPs with infinite time-horizon. Here two points are discussed:
\begin{enumerate}[(1).]
\item In the perspective of probability theory, MDP models with absorbing states to maximize the expected returns generalize infinite-horizon MDPs that maximize discounted expected return. To see it, consider an infinite-horizon discounted MDP $M= (\mathcal{S}_0, \mathcal{A}, r, {P}, \mu,\gamma)$.
To create a new model $\tilde{M}=({\cal S},{\cal A},\tilde{r},\tilde{P},\mu)$ with an absorbing state, we introduce a new state $\xi$ such that ${\cal S}={\cal S}_0\cup\{\xi\}$, define $$
\tilde P(s{'}|s,a)=\left\{\begin{array}{lr}
\gamma {P}(s{'}|s,a) & \text { if } s, s{'} \in \mathcal{S}_0 \\
1-\gamma & \text { if } s \in \mathcal{S}_0, s{'}=\xi \\
1 & \text { if } s=s{'}=\xi ,
\end{array}\right.
$$
keep the action space the same, and introduce a reward function $\tilde{r}(s,a)$ by $\left\{\begin{array}{ll}
r(s,a)&\hbox{if }s\in {\cal S}\\
0&\hbox{otherwise}\end{array}\right.$.  The optimal policy of $M$ and $\tilde{M}$ are the same because the two models share the same Bellman optimality equation(a similar argument is discussed in Section 10.1 of \cite{altman1999constrained}).

 In the context of learning, $M$ and $\tilde{M}$ are essentially two different models because one generally needs to estimate the unknown parameter $\gamma$ (the probability of transitioning to the absorbing state or the unknown discount factor) in $\tilde M$, whereas $\gamma$ is known in $M$.
\item In general, an MDP with absorbing states to maximize expected return is not necessarily translated to an MDP to maximize expected discounted return. For example, one can consider the case when the transition probability of a state to the absorbing state under a policy depends on the state.
\end{enumerate}
\end{remark}}


\section{Minimax Weight Learning for absorbing MDP (MWLA)}\label{section3}

Let
\begin{equation}\label{behavior trajectories}
Z=(s_0, a_0, s_1, a_1,\dots, s_{T-1}, a_{T-1})\hbox{ and }Z_i=(s_0^i, a_0^i, s_1^i, a_1^i,\dots, s_{T_i-1}^i, a_{T_i-1}^i), i=1,2,\dots, m
\end{equation}
 be a representative episode of an absorbing MDP with probability distribution $P_{\mu,\pi_b}$ and its i.i.d. copies, respectively. The objective is to estimate the expected return $R_{\pi_e}$ by using the dataset $D$ of $m$ i.i.d. trajectories $\{\boldsymbol{\tau}^{i}, i=1,\cdots, m\}$, each of which is a realization of $Z_i$ truncated at a prespecified time step $H$, i.e.
\beqnn
\boldsymbol{\tau}^{i}=\left(s_{0}^{i}, a_{0}^{i}, r_{0}^{i}, s_{1}^{i}, a_{1}^{i}, r_{1}^{i}, \dots, s_{T_{i}\wedge H}^{i}, a_{T_{i}\wedge H-1}^{i}, r_{T_{i}\wedge H-1}^{i},s_{T_{i}\wedge H}^{i}\right), i=1,2,\dots,m.\eeqnn
The algorithm developed here, as indicated by the title ``Minimax Weight Learning for Absorbing MDP'' of this section and referred to as MWLA below is an extension of \cite{uehara2020minimax}'s Minimax Weight Learning (MWL) algorithm for infinite-horizon discounted MDPs.
Their method uses a discriminator function class $\mathcal{Q}$ to learn the importance weight $w$ (defined in equation (\ref{defw2}) below) on state-action pairs. One of their important tools is the (normalized) discounted occupancy, which can be approximated well using the given discount factor $\gamma$ and the suitable dataset (for example, the dataset consisting of i.i.d. tuples $(s, a, r, s')$).
{However, in our setting, the normalized occupancy is not applicable since the reward is not discounted.}
Our method is essentially based on the occupancy measure defined in equation (\ref{a1}).

For any $(s, a)\in \S_0\times\A$, let
\beqlb\label{defw2}
w_{\frac{\pi_e}{\pi_{b}}}(s,a)\dot=\frac{d_{{\pi_e}}(s,a)}{d_{\pi_{b}}(s,a)},  \hbox{a.e. } d_{\pi_{b}}.
\eeqlb
Observe that
$$
R_{\pi_e}=\sum_{(s,a)\in S_0\times\A}R(s,a)d_{\pi_e}(s,a)=\sum_{(s, a)\in\S_0\times\A}w_{\frac{\pi_e}{\pi_b}}(s,a)R(s,a)d_{\pi_b}(s,a) =\Phi_{\pi_b}(w_{\frac{\pi_e}{\pi_b}}R),
$$
provided that $d_{\pi_e}(s, a)>0$ implies $d_{\pi_b}(s, a)>0$.
By replacing $d_{\pi_b}$, $R$ and $w_{\frac{\pi_e}{\pi_b}}$ with their estimates $\hat d_{\pi_b}$, $\hat{R}$ and $\hat w_{\frac{\pi_e}{\pi_b}}$, respectively, a plug-in approach suggests that $R_{\pi_e}$ can be simply estimated by
 \begin{equation}\label{EstimatingReturn}
 \hat R_{\pi_e}=\hat{\Phi}_{\pi_b}(\hat w_{\pi_e\over\pi_b}\hat{R})\doteq\sum_{(s, a)\in\S_0\times\A}\hat w_{\frac{\pi_e}{\pi_b}}(s,a)\hat{R}(s,a)\hat d_{\pi_b}(s,a),
 \end{equation}
in which the crucial step is to estimate $w_{\frac{\pi_e}{\pi_b}}$.

From equality \eqref{f4}, we formally introduce an error function
 \beqlb\label{defL}
L(w,q)&\doteq &{\displaystyle\sum_{(s,a,s')\in\S_0\times\A\times\S_0}}w(s,a)q(s',{\pi_e})d_{{\pi_b}}(s,a,s')\notag\\
&&-{\displaystyle\sum_{(s,a)\in\S_0\times\A}}w(s,a)q(s,a)d_{{\pi_b}}(s,a)+\mathrm{E}_{s\sim \mu}\big[q(s,{\pi_e})\big],
\eeqlb
so that
$ L(w_{\frac{\pi_e}{\pi_b}}, q)=0\hbox{ for all }q\in \Xi.$
Conversely, by taking a family of particular functions $\{q(s,a)={\bm 1}_{(\bar s,\bar a)}(s, a): (\bar s,\bar a)\in\mathcal {S}_0\times\mathcal{A}\}$, as what has been done  in \eqref{a1}, the uniqueness of the solution to this system of equations can be derived, as what is in the following Theorem.

Note that \(L(w_{\frac{\pi_e}{\pi_b}}, q)=0\hbox{ for all }q\in \Xi\) if and only if \(L(w_{\frac{\pi_e}{\pi_b}}, q)=0\hbox{ for all }q\in \Xi, \|q\|_2^2= 1\), we can safely work with the system of equations $L(w,q)=0,\|q\|_2^2= 1$, where and throughout the remainder of the paper, $\|\cdot\|_2$ of a vector or a matrix denotes its Eclidean norm.

\begin{theorem}\label{Theorem 1} The function $w=w_{\frac{\pi_e}{\pi_b}}$ is the unique bounded solution to the system of equations \(\{L(w,q)=0: \|q\|_2^2= 1\}\), provided that the following conditions hold:
\begin{enumerate}[1)]
\item The occupancy measure $d_{{\pi_e}}$ is the unique solution to the system of equations of $q$,
\beqlb\label{U2}
q(s, a)=\mu(s)\pi_e(a|s)+\sum_{(s',a')\in \S_0\times\A}P(s|s',a')\pi_e(s|a)q(s', a'), \;\; (s, a)\in\S_0\times\A.
\eeqlb
\item $d_{\pi_e}(s,a)>0$ implies $d_{\pi_{b}}(s,a)>0$ for all $(s,a)\in S_0\times \A$.
\end{enumerate}
\end{theorem}
Theorem \ref{Theorem 1} simply states that
\begin{equation}\label{f5-1}
w_{\frac{\pi_e}{\pi_b}}=\operatorname*{argmin}\limits_{ w}\max\limits_{\|q\|_2^2= 1 }L(w, q)^{2}.
\end{equation}
In real applications, $w_{\frac{\pi_e}{\pi_b}}$ is approximated by
\beqnn
{w}^*(s,a)\doteq\operatorname*{argmin}\limits_{ w\in \mathcal{W}}\max\limits_{q \in \mathcal{Q}}L(w, q)^{2},
\eeqnn
where $\mathcal{W}\subset \Xi$ is a working class of $w_{\frac{\pi_e}{\pi_b}}$, and $\mathcal{Q}\subset\Xi$ treated as discriminators. For artificially designed $\cal Q$, the class $\cal Q$ is only required to be bounded because it may be computationally inefficient to require $\|q\|_2^2= 1$ for $q\in \cal Q$.

For any $(s,a)\in\S_0\times\A$, define a function $V_{s,a,\pi_e}\in \Xi$ by \(
V_{s,a,\pi_e}(s',a')\doteq\sum_{t=0}^\infty \P_{(s, a),{\pi_e}}(s_t=s',a_t=a')\). The theorem below, with a novel  relationship between $w_{\pi_e\over\pi_b}-w$ and the error function $L$, will be helpful in bounding the estimation error of occupancy measure ratio by means of the mini-max loss via the discriminator class $\mathcal{Q}$ chosen properly.

\begin{theorem}\label{Theorem 2}The error function allows for the expressions
\begin{equation}\label{specialError}
 L\left(w,V_{s',a',\pi_e}\right)=d_{\pi_e}(s',a')-w(s',a')d_{\pi_{b}}(s',a') \hbox{ and } L\left(w,V_{s',a',\pi_e}/d_{\pi_{b}}(s',a')\right)=w_{\pi_e\over\pi_b}-w.
\end{equation}
Consequently,
\begin{enumerate}[(1)]
\item $ \| d_{\pi_e}-w d_{\pi_{b}}\|_{\infty}\leq\max\limits_{q\in\mathcal{Q}}|L(w,q)|
\hbox{ if } \{\pm V_{s',a',\pi_e}: (s',a')\in\mathcal{S}_0\times\mathcal{A}\}\subseteq \mathcal{Q}
$ and
\item $ \| w_{\frac{\pi_e}{\pi_b}}-w^*\|_{\infty}\leq \min\limits_{w\in\mathcal{W}}\max\limits_{q\in\mathcal{Q}}|L(w,q)|
\hbox { if }\{\pm V_{s',a',\pi_e}/d_{\pi_{b}}(s',a'):(s',a')\in\mathcal{S}_0\times\mathcal{A}\}\subseteq \mathcal{Q}
$.
\end{enumerate}
Here $\|\cdot\|_\infty$ denotes the supremum norm.
\end{theorem}

The MWLA algorithm to estimate ${R}_{\pi_e}$ is now ready to be described as follows.
Firstly, 
for all $(s,a,s')\in\S_0\times\A\times\S_0$, define the trajectory-specific empirical occupancy measures and  rewards by
 \beqlb\label{defUb}
 \hat{d}^{i}_{\pi_{b}}(s,a)\doteq\sum_{t=0}^{l_i-1}{\bm 1}_{(s, a)}(s_t^i, a_t^i),\;~ \hat{d}^{i}_{\pi_{b}}(s,a, s')\doteq\sum_{t=0}^{l_i-1}{\bm 1}_{(s, a, s')}(s_t^i, a_t^i, s_{t+1}^i)
 \eeqlb
and
 \beqnn
  \hat{R}^i(s, a)\doteq\frac{\sum_{t=0}^{l_i-1}r_t^{i}{\bm 1}_{(s, a)}(s_t^i, a_t^i)}{\hat{d}^{i}_{\pi_{b}}(s,a)} \hbox{ if } {\hat{d}^{i}_{\pi_{b}}(s,a)}>0 \hbox{ and } 0\;\;\hbox{otherwise}.
  \eeqnn
Secondly,
%
for any $w, q\in B(\S_0\times\A)$, introduce an empirical error function
 \beqlb\label{f2}
\hat{L}_{m}(w,q)&=&\frac{1}{m}\sum_{i=1}^{m}\sum_{(s,a,s')\in\S_0\times\A\times\S_0} w(s,a)q(s',{\pi_e})\hat{d}^{i}_{\pi_{b}}(s,a,s')\nonumber
\\&-&\frac{1}{m}\sum_{i=1}^{m}\sum_{(s,a)\in\S_0\times\A}w(s, a)q(s,a)\hat{d}^{i}_{\pi_{b}}(s,a)+\mathrm{E}_{s\sim\mu}\big[q(s,{\pi_e})\big],
\eeqlb
so that an estimate of $w_{\frac{\pi_e}{\pi_b}}$ is then defined by
 \beqnn
 \hat{w}_{m}(s,a)\doteq\operatorname*{argmin}\limits_{ w\in \mathcal{W}}\max\limits_{q \in \mathcal{Q}} \hat{L}_{m}(w, q)^{2}.
 \eeqnn
Putting them into equation \eqref{EstimatingReturn}, the expected return $R_{\pi_e}$ is consequently estimated by
\begin{equation}\label{EstimatedReturn}
\hat{R}_{\pi_e,m}=\frac{1}{m}\sum\limits_{i=1}^{m}\sum_{(s,a)\in\S_0\times\A}\hat{w}_m(s,a)\hat{R}^i(s,a)\hat{d}^{i}_{\pi_{b}}(s,a).
\end{equation}
 \begin{remark}\label{rem3.1}
Compared to the MWL algorithm in \cite{uehara2020minimax}, the MWLA algorithm described above utilizes truncated episodes instead of the $(s, a, r, s')$ tuples. Consequently, the accuracy of the estimation depends on two factors: the data size $m$ and the truncation level $H$. Therefore, it is crucial to comprehend how errors vary at different levels of $m$ and $H$. This will aid us in gaining a better understanding of the effects of episode truncation and identifying an appropriate truncation level $H$ under which the errors caused by truncation can be deemed acceptable. A detailed analysis on this matter is presented in the next section.
\end{remark}
Below is an example of the MWLA algorithm applied to absorbing tabular MDPs.
\begin{example}\label{ex3.1}
{\rm Write $\S_0=\{0,1,\dots,n-1\}\) and \(\A=\{0,1,\dots,h-1\}$ for the tabular model. Note that any matrix \({\bm u}\dot=(u_{kl})\in \mathbb{R}^{n\times h}$ defines a map ${\bm u}:\S_0\times\A\mapsto\mathbb{R}$ by ${\bm u}(k,l)=u_{kl}$.
Take the function classes \(\mathcal{W}=\mathbb{R}^{n\times h}$ and $\mathcal{Q}=\big\{{\bm u}\in \mathbb{R}^{n\times h}: \|\bm u\|_2^2= 1\big\}\).
For every $0\leq k\leq n-1, 0\leq l\leq h-1$, denote by ${\bf 1}_{(k,l)}$ the $nh$-dimensional unit column vector with $1$ at its $(kh+l)$-th component, and let ${\bf 1}_{(k,\pi_e)}=\sum_{l=0}^{h-1}\pi_e(l|k){\bf 1}_{(k,l)}$. For and ${\bm u}\in {\cal W}$ and ${\bm v}\in{\cal Q}$, the empirical error is
\beqnn
 \hat{L}_{m}({\bm u},{\bm v})&=&\frac{1}{m}\sum_{i=1}^m\sum_{(k,l,k')\in\S_0\times\A\times\S_0}{\bm u}(k,l){\bm v}(k',\pi_e)\hat{d}^i_{\pi_b}(k,l,k')
 \\&& -\frac{1}{m}\sum_{i=1}^m\sum_{(k,l)\in\S_0\times\A}{\bm v}(k,l)\hat{d}^i_{\pi_b}(k,l)
 +\sum_{k\in \S_0}\mu(k){\bm v}(k,\pi_e)
  \\&=&(\overset{\rightarrow}{\bm u}^{\top}{\bm\hat{A}}+{\bm b}^{\top})\overset{\rightarrow}{\bm v},
  \eeqnn
 where
  \beqnn
  {\bm\hat{A}}&=&\frac{1}{m}\sum_{i=1}^{m}\sum_{k=0}^{n-1}\sum_{l=0}^{h-1}\sum_{v=0}^{n-1}\mathbf{1}_{(k,l)}\mathbf{1}_{(v,\pi_e)}^{\top}
  \hat{d}^{i}_{\pi_{b}}(k,l,v)-\frac{1}{m}\sum_{i=1}^{m}\sum_{k=0}^{n-1}\sum_{l=0}^{h-1} \mathbf{1}_{(k,l)}\mathbf{1}_{(k,l)}^{\top}
    \hat{d}^{i}_{\pi_{b}}(k,l)
  \\&=&\frac{1}{m}\sum_{i=1}^{m}\sum_{t=0}^{T_i\wedge H-1}{\bf 1}_{(s^i_t, a^i_t)}\Big[\sum_{a\in\A}\pi_e(a|s^i_{t+1}){\bf 1}^{\top}_{(s_{t+1}^i, a)}-{\bf 1}^{\top}_{(s_t^i, a^i_t)}\Big]
  \eeqnn
 is an $nh\times nh$ matrix, $\overset{\rightarrow}{\bm u}$ is the vectorized ${\bm u}$ by columns and ${\bm b}=\sum_{(s,a)\in\S_0\times\A}\mu(s)\pi_e(a|s){\bf 1}_{(s,a)}$ is an $nh$-vector. Therefore,
   $$\max_{{\bm v}:\|{\bm v}\|^2= 1}\hat{L}^2_{m}({\bm u},{\bm v})=\max_{{\bm v}:\|{\bm v}\|^2= 1}\overset{\rightarrow}{\bm v}^{\top}(\bm\hat{A}^{\top}\overset{\rightarrow}{\bm u}+{\bm b})(\bm\hat{A}^{\top}\overset{\rightarrow}{\bm u}+{\bm b})^{\top}\overset{\rightarrow}{\bm v}={\|\bm\hat{A}^{\top}\overset{\rightarrow}{\bm u}+{\bm b}\|}_2^2.$$
Therefore, the estimate is $\hat{w}_m={{\hat {\bm u}}}$ with $\overset{\rightarrow}{\hat{\bm u}}$ the least square solution to the equation \({\bm\hat{A}^{\top}} \overset{\rightarrow}{\bm u}=-{\bm b}
\).
}
\end{example}
\begin{remark}[\bf A variant for known $\pi_b$]\label{MSWLA}\rm If we define $d_\pi(s)=\Phi_\pi({\bf 1}_{\{s\}})$, then $d_\pi(s, a)=d_{\pi}(s)\pi(a|s)$ and from \eqref{f4}, we have that
 \beqnn
\sum_{(s,a,s')\in\S_0\times\A\times\S_0}q(s',{\pi})d_{{\pi}}(s)\pi(a|s)P(s'|s,a)-\sum_{s\in\S_0}q(s,\pi)d_{{\pi}}(s)+\mathrm{E}_{s\sim \mu}\big[q(s,{\pi})\big]=0.
\eeqnn
For a given target policy $\pi_e$, simply denote $q(s,\pi_e)$ by $q(s)$, so that the equation above can be rewritten as
 \beqnn
\sum_{(s,a,s')\in\S_0\times\A\times\S_0}w_{\frac{\pi_e}{\pi_b}}(s)q(s'){\pi_e(a|s)\over \pi_b(a|s)}d_{{\pi}_b}(s,a,s')-\sum_{s\in\S_0}w_{\frac{\pi_e}{\pi_b}}(s)q(s)d_{{\pi}_b}(s)+\mathrm{E}_{s\sim \mu}\big[q(s)\big]=0,
\eeqnn
where  $w_{\frac{\pi_e}{\pi_b}}(s)={d_{\pi_e}(s)\over d_{\pi_b}(s)}$.

With this equation, if the behavior policy $\pi_{b}$ is known, we can construct a corresponding estimate of the value function based on the minimax optimization problem:
\begin{equation*}
\min\limits_{w \in \mathcal{W}^{s}} \max\limits_{q\in \mathcal{Q}^{s}}\big(\sum_{(s,a,s')\in\S_0\times\A\times\S_0}w_(s)q(s'){\pi_e(a|s)\over \pi_b(a|s)}d_{{\pi}_b}(s,a,s')-\sum_{s\in\S_0}w(s)q(s)d_{\pi_b}(s)+\mathrm{E}_{s\sim \mu}\big[q(s)\big]\big)^{2}.
\end{equation*}
For convenience, we refer to the method as MSWLA (marginalized state weight learning for absorbing MDPs) algorithm which is essentially an extension of the method discussed in \cite{liu2018breaking}.
By similar arguments in Example \ref{ex3.1}, in the tabular case
 where $\S_0=\{0,1,\dots,n-1\}$ and the function classes $\mathcal{W}^s$ and $\mathcal{Q}^s$ are $\mathbb{R}^{n}$,
the empirical error function for the MSWLA algorithm is $\hat{L}_{m}({\bm u},{\bm v})=({\bm u}^{\top}{\bm\hat{A}}+{\bm b}^{\top}){\bm v}$ for any $\bm u\in \mathcal{W}^s, \bm v\in \mathcal{Q}^s$,
 where
  \beqnn
  {\bm\hat{A}}&=&
\frac{1}{m}\sum_{i=1}^{m}\sum_{t=0}^{T_i\wedge H-1}{\bf 1}_{\{s^i_t\}}\Big[{\pi_e(a^i_t|s^i_t)\over \pi_b(a^i_t|s^i_t)}{\bf 1}^{\top}_{\{s_{t+1}^i\}}-{\bf 1}^{\top}_{\{s_t^i\}}\Big],
  \eeqnn
${\bm b}=\sum_{s\in\S_0}\mu(s){\bf 1}_{\{s\}}$, and for any $s\in\S_0$, ${\bf 1}_{\{s\}}$ is the $n$-dimensional column vector whose $s$-th entry is $1$ and other elements are $0$.
\end{remark}

\section{ MSE bound of the estimated return}\label{section4}

Denote by
 \(Q_{\pi_e}(s, a)\doteq\E_{(s,a),\pi_e}(\sum_{t=0}^{T-1}r(s_{t}, a_{t}))
\)
the commonly known Q-function and $H_m $ the unique positive solution to the equation $2mx^2+2\ln m\ln x-\ln m=0.$ Let us now analyze the error bound of $\hat{R}_{\pi_e,m}$ with respect to the number of episodes $m$ and the truncation level $H$, measured by the mean squared error (MSE)
as provided in the following theorems.

The following technical assumption is necessary, which is also supposed in \cite{uehara2020minimax} as Assumption 2.

 \begin{assumption}\label{Assumption 3}
There exists a constant $C_w>0 $, such that $\sup_{(s, a) \in \S_0\times \A}w_{\frac{\pi_e}{\pi_b}}(s,a) \leq C_w$.
\end{assumption}

\begin{theorem}\label{Theorem 3}  Suppose that
\begin{enumerate}[1)]
 \item there exists a common envelop $G$ of $\cal W$ and $\cal Q$, i.e. $|w|\leq G$, $|q|\leq G$ for all $w\in\mathcal{W}$, $q\in\mathcal{Q}$, satisfying
     \beqlb\label{Assumption4-2}
     \Lambda_1:=\sum_{(s, a)\in\S_0\times \A} G(s, a)<+\infty \;\;\text{and}\;\; \Lambda_2:=\sum_{(s, a)\in\S_0\times \A} G^2(s, a)<+\infty;
     \eeqlb
 \item $\mathcal{W}$ and $\mathcal{Q}$ have finite pseudo-dimensions $D_{\mathcal{W}}$ and $D_{\mathcal{Q}}$, respectively;
 \item Assumptions \ref{Assumption 1} and \ref{Assumption 3} hold;

  \item $Q_{\pi_e}\in\mathcal{Q}$;

 \item there exists $\lambda_0>0$ such that $\mathrm{E}_{\mu,\pi_b}(e^{\lambda_0 T})=M_0<\infty$.
   \end{enumerate}
Then we have the following:
   \begin{enumerate}[1)]
\item When $M_0e^{-\lambda_0 H}> H_m$, there exists a constant $C$ independent of $H$ and $m$, such that
 \beqnn
    \mathrm{E}\big[(\hat{R}_{\pi_e,m}-R_{{\pi_e}})^{2}\big]&\leq& C\Big(e^{-2\lambda_0 H}+\frac{H^2\ln m}{m}\Big)+8\min\limits_{w\in\mathcal{W}}\max\limits_{q\in\mathcal{Q}}L(w,q)^{2}.
\eeqnn
\item When $M_0e^{-\lambda_0 H}\leq H_m$, there exists a constant $C$ independent of $H$ and $m$, such that
\beqnn
    \mathrm{E}\big[(\hat{R}_{\pi_e,m}-R_{{\pi_e}})^{2}\big]&\leq& C\frac{\ln^3 m}{m}+8\min\limits_{w\in\mathcal{W}}\max\limits_{q\in\mathcal{Q}}L(w,q)^{2}.
\eeqnn
Especially, if $w_{\frac{\pi_e}{\pi_b}}\in\mathcal{W}$ and $M_0e^{-\lambda_0 H}\leq H_m$,  then
$$\mathrm{E}\big[(\hat{R}_{\pi_e,m}-R_{{\pi_e}})^{2}\big]\leq C{\frac{\ln^3 m}{m}}.$$
\end{enumerate}
\end{theorem}

\begin{theorem}\label{Corollary 4} Suppose the assumptions in Theorem \ref{Theorem 3} hold and $m\geq e$ but $Q_{\pi_e}\not\in\mathcal{Q}$.
\begin{itemize}
\item[(1)] When $M_0e^{-\lambda_0 H}> H_m$, there exists a constant $C$ independent of $H,m$, such that
 \beqnn
    \mathrm{E}\big[(\hat{R}_{\pi_e,m}-R_{{\pi_e}})^{2}\big]&\leq& C\Big(e^{-2\lambda_0 H}+\frac{H^2\ln m}{m}\Big)+16\min\limits_{w\in\mathcal{W}}\max\limits_{q\in\mathcal{Q}}L(w,q)^{2}\\
    &&+4\max\limits_{w\in\mathcal{W}}\min\limits_{q\in\mathcal{Q}}L^2(w,Q_{\pi_e}-q).
\eeqnn
\item[(2)] When $M_0e^{-\lambda_0 H}\leq H_m$, there exists a constant $C$ independent of $H,m$, such that
\beqnn
    \mathrm{E}\big[(\hat{R}_{\pi_e,m}-R_{{\pi_e}})^{2}\big]&\leq& C\frac{\ln^3 m}{m}+16\min\limits_{w\in\mathcal{W}}\max\limits_{q\in\mathcal{Q}}L(w,q)^{2}+4\max\limits_{w\in\mathcal{W}}\min\limits_{q\in\mathcal{Q}}L^2(w,Q_{\pi_e}-q).
\eeqnn
\end{itemize}
Obviously, the additional term
$\max\limits_{w\in\mathcal{W}}\min\limits_{q\in\mathcal{Q}}L(w, Q_{{\pi_e}}-q)^{2}$ becomes $0$ when
$Q_{{\pi_e}}$ in the closure of $\mathcal{Q}$ under the metric $||\cdot||_\infty$.
\end{theorem}

Theorems \ref{Theorem 3} and \ref{Corollary 4} provide upper bounds for the MSE of the MWLA algorithm for a few situations. They are expressed as functions of two key parameters: the truncation level $H$ and the number of the episodes $m$. When $H$ is small (i.e., $M_0e^{-\lambda_0 H}> H_m$), the estimate errors composed of four parts: the pure truncation term $e^{-2\lambda_0 H}$,  a crossing term $H^2\ln m/m$ arising from the sampling randomness, an approximation error $\min\limits_{w\in\mathcal{W}}\max\limits_{q\in\mathcal{Q}}L^2(w,q)$, and an optimization error $\max\limits_{w\in\mathcal{W}}\min\limits_{q\in\mathcal{Q}}L^2(w,Q_{\pi_e}-q)$. While  the first two errors stem from the randomness of the statistics, the other two result from the degree of closeness of the two function classes $\mathcal{W}$ and $\mathcal{Q}$ to $w_{{\pi}/\pi_b}$ and $Q_{{\pi}}$, respectively. For a large $H$ (i.e., $M_0e^{-\lambda_0 H}\leq H_m$), however, the pure truncation term $e^{-2\lambda_0 H}$ and the mixing term $H^2\ln m/m$ can be dominated by an $H$-free term $C\ln^3 m/m$.  This indicates simply that MWLA algorithm can avoid the curse of the horizon.

In the following are more remarks on the results.

\begin{remark} Consider the case $Q_{\pi_e}\in\mathcal{Q}$. For the infinite horizon MDP with $m$ i.i.d. tuples $(s_i,a_i,s_i',r_i)$, the error bound of the MWL method consists of a statistical error $\frac{\ln m}{m}+\mathcal{R}_{m}^{2}(\mathcal{W},\mathcal{Q})$ and an approximation error
$\min\limits_{w\in\mathcal{W}}\max\limits_{q\in\mathcal{Q}}L(w,q)^{2}$,
where $\mathcal{R}_{m}(\mathcal{W},\mathcal{Q})$ is the Rademacher complexity of the function class $$\left\{\left(s, a, s{'}\right) \mapsto |w(s, a)(q(s^{'}, \pi)-q(s, a))|: w \in \mathcal{W}, q \in \mathcal{Q}\right\},$$ as given in Theorem 9 of \cite{uehara2020minimax}.
Let $D_{\mathcal{W}}$,$D_{\mathcal{Q}}$ be the VC-subgraph dimension (i.e. pseudo-dimension) of $\mathcal{W},\mathcal {Q},$ respectively. Because $\mathcal{R}_{m}(\mathcal{W},\mathcal{Q})=O(\sqrt{\frac{\max{(D_{\mathcal{W}},D_{\mathcal{Q}})}}{m}})$ (Corollary 1 of \citealp{uehara2021finite}), the statistical error is dominated by $\frac{\ln m}{m}$.  For MWLA with $m$ i.i.d. episodes, the MSE bound also consists of an approximation error $\min\limits_{w\in\mathcal{W}}\max\limits_{q\in\mathcal{Q}}L(w,q)^{2}$ and a statistical error. When $M_0e^{-\lambda_0 H}\leq H_m$, the statistical error is bounded by $\frac{\ln^3 m}{m}$, including an extra factor $\ln^2 m$ in form. 
\end{remark}

\begin{remark} For $m>e$, one has $\sqrt{\ln m/(2m)}<H_m<\ln m\sqrt{e/m}$ (Lemma \ref{Lemma 11} in Appendix). Therefore, when $M_0e^{-\lambda_0 H}> H_m$, it follows that $H\leq \ln M_0+\ln 2/2+\ln m/2$ and $\frac{H^2\ln m}{m}\leq C\frac{\ln ^3m}{m}$ for some constant $C$. Whatever $H$ is, the bounds in Theorem \ref{Theorem 3} are both less than $$C(e^{-2\lambda_0H}+\ln^3 m/m)+8\min\limits_{w\in\mathcal{W}}\max\limits_{q\in\mathcal{Q}}L(w,q)^{2}.$$
\end{remark}

\begin{remark}\label{rem4.4} In the tabular setting, if we take $\mathcal{W}=\{w: \|w\|_2\leq K_0\}$ and $\mathcal{Q}=\{q: \|q\|_\infty<K_1\}$, where $K_0$ is a constant larger than $C_w$ in Assumption \ref{Assumption 3}, then all assumptions in Theorem \ref{Theorem 3} hold. Hence,
$$\mathrm{E}\big[(\hat{R}_{\pi_e,m}-R_{{\pi_e}})^{2}\big]\leq C\Big(e^{-2\lambda_0 H}+\frac{\ln^3 m}{m}\Big).$$
\end{remark}

{\begin{remark}\label{rem4.5}
With an MSE bound, a confidence interval of the error of the estimation can be derived easily by Markov's inequality. That is, if $Q_{\pi_e}\in\mathcal{Q}$ and $w_{\frac{\pi_e}{\pi_b}}\in\mathcal{W}$, then
\beqnn
P(|\hat{R}_{\pi_e,m}-R_{\pi_e}|>\epsilon)<\frac{\E((\hat{R}_{\pi_e,m}-R_{{\pi_e}})^{2})}{\epsilon^2}\leq\frac{C\Big(e^{-2\lambda_0 H}+\frac{\ln^3 m}{m}\Big)}{\epsilon^2}.
\eeqnn
As a result, for any given $\delta$ and $\epsilon$, one can easily retrieve a sample complexity $ m(\epsilon,\delta,H)$ or $ H(\epsilon,\delta,m)$, such that $P(|\hat{R}_{\pi_e,m}-R_{\pi_e}|>\epsilon)<\delta$ if $m>m(\epsilon,\delta,H)$ or $H>H(\epsilon,\delta,m)$.

\end{remark}
}

\section{ Connections to related work}\label{section5}

Research on off-policy evaluation (OPE) for MDPs with infinite and fixed finite horizons can be classified into two categories according to whether the behavior policy is known.

When the behavior policy is known, IS is a commonly used method that reweights rewards obtained by behavior policies, according to its likelihood ratio of the evaluation policy $\pi_e$ over the behavior $\pi_{b}$ to provide unbiased estimates of the expected returns. However, the IS method suffers from exponentially increasing variance over the time horizon because the ratio is computed as a cumulative product of the importance weight over action $\frac{\pi_e(a|s)}{\pi_{b}(a|s)} $ at each time step \citep{precup2000eligibility}. To reduce that extremely high variance, a series of OPE methods have been proposed based on IS. For example, the weighted importance sampling (WIS) method, the stepwise importance sampling method, and the doubly robust(DR) method can reduce the variance to certain degree (\citealp{cassel1976some}; \citealp{robins1994estimation}; \citealp{robins1995semiparametric}; \citealp{bang2005doubly}). However, the exponential variance of IS-based methods cannot be significantly improved when the MDP has high stochasticity \citep{jiang2016doubly}.

The MIS method proves a promising improvement over IS by successfully avoiding the trouble of exponential variance. For example, for a finite-horizon inhomogeneous MDP, compared to weighting the whole trajectories, \cite{xie2019towards} uses a ratio $w_{t}(s)\frac{\pi_e(a|s)}{\pi_{b}(a|s)}$ with $w_{t}(s)=\frac{d_{\pi_e,t}(s)}{d_{\pi_{b},t}(s)}$ to reweight the rewards $r$
 in order to achieve a lower variance. In an infinite horizon setting, based on a discounted stationary distribution, \cite{liu2018breaking} proposes using the ratio $w_{\pi_e\over\pi_{b}}(s)\cdot \frac{\pi_e(a|s)}{\pi_{b}(a|s)} $ with $w_{\pi_e\over\pi_{b}}(s)=\frac{d_{\pi_e,\gamma}(s)}{d_{\pi_{b},\gamma}(s)}$. The ratio $w_{\pi_e\over\pi_{b}}(s)$ is then estimated by a minimax procedure with two function approximators: one to model a weight function $w_{\pi_e\over\pi_{b}}(s)$, and the other to model $V^{\pi_{b}}$, as a discriminator class for distribution learning.

For the case of unknown behavior policies, \cite{hanna2019importance} show that the IS method with an estimated behavior policy has a lower asymptotic variance than the one with a known behavior strategy. The fitted $Q$-iteration, which uses dynamic programming to fit $Q_{\pi_e}$ directly from the data, can overcome the curse of dimensionality, with a cost of assuming that the function class contains $Q_{\pi_e}$ and is closed under the Bellman update $B^{\pi_e}$, so as to avoid a high bias, see \cite{ernst2005tree} and \cite{le2019batch}. \cite{uehara2020minimax} propose the MWL algorithm by estimating marginalized importance weight \(
w_{\pi_e\over\pi_{b}}(s,a)=\frac{d_{\pi_e,\gamma}(s,a)}{d_{\pi_{b}}(s,a)}\). A Dualdice algorithm is further proposed to estimate the discounted stationary distribution ratios
(\citealp{nachum2019dualdice}; \citealp{nachum2019algaedice}; \citealp{nachum2020reinforcement}) where the error function can be considered as a derivative of the error function (loss function in their terminology) in \cite{uehara2020minimax}. \cite{jiang2020minimax} combine MWL and MQL into a unified value interval with a unique type of double robustness, if either the value-function or the importance-weight class is correctly specified, the interval is valid, and its length measures the misspecification of the other class.

In reinforcement learning, while many benchmark environments are indeed episodic and have random horizons, such as board games (a game terminates once the winner is determined), trips through a maze, and dialog systems (a session terminates when the conversation is concluded) (\citealp{jiang2017theory}), there are only limited efforts specifically contributed to absorbing MDPs. Researchers often take absorbing MDPs as special cases of finite-horizon MDPs, by padding all trajectories with absorbing states (with random lengths) to the same length. Another way to handle absorption practically is to use the infinite-horizon setup (with a sufficiently large discount factor), and whenever a trajectory terminates, we imagine it continuous infinitely at absorbing states.
However, when the random horizons are not bounded and the random episodes are not observed completely, especially, accompanied by the undiscounted rewards, new issues will arise. For example, how do the unobserved trajectories affect the results? As our results show this problem is by no means trivial, which is essentially neglected when we simply apply the two ways mentioned above.

The current paper deals with the OPE for absorbing MDPs through the MWLA algorithm, a variant of the MWL to fit the random horizon and truncated episodic data modeled by absorbing MDPs, using episodic data rather than the  $(s, a, r, s')$-tuple data. In addition, we explicitly analyze the dependence of the error bound of the MSE on the truncation level and data size and derive the uniform bound of the MSE by optimization when the truncation level is relatively large.


\section{Experiments}\label{section6}

In this section, we present several computational experiments that showcase the performance of the MWLA and other relevant algorithms. We first describe the experimental settings and subsequently report and discuss the results.

\subsection{Setting}
\begin{figure}[htbp]
\centering
\includegraphics[width=0.15\textwidth]{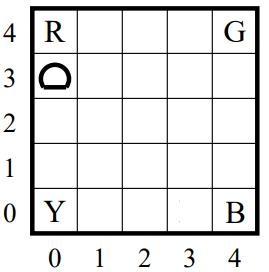}
\caption{Taxi Grid}
\label{Taxi Grid}
\end{figure}

The environment we employ is a version of \cite{dietterich2000hierarchical}'s Taxi, a two-dimensional setup that simulates a taxi moving along a $5\times 5$ grid world, as indicated by Figure \ref{Taxi Grid}. The four corners are marked as R(ed), B(lue), G(reen), and Y(ellow).  Initially, the taxi randomly chooses a corner to wait for a passenger, who appears or disappears with a probability at each of the four corners, and that passenger wishes to be transported to one of the four corners (also chosen randomly). The taxi must pick up the passenger and drops him off at his destination. An episode ends once a passenger is successfully dropped off at his destination.

There are a total of 2000 states ($25\times2^{4}\times5$), made from 25 taxi locations, $2^{4}$ passenger appearance status, and 5 taxi status (empty or one of 4 destinations with a passenger). There are four navigation actions that move the taxi one square North, South, East, or West, respectively. Each action yields a deterministic movement. Only 3 and 2 actions can be taken when the taxi is at the boundary and the corner, respectively. The taxi receives a reward of -1 at every time step when a passenger is picked up in Taxi, 0 if the passenger is successfully dropped off at the right place and -2 at each time step if the taxi is empty.

 As in  \cite{liu2018breaking}, we first run a Q-learning with a soft-max action preference with 400,000 iterations to produce the target policy $\pi_e$ and then run 60,000 iterations to produce another auxiliary policy $\pi^+$, of which both are regularized such that the probability of crossing any boundary of the grid is $0$. The behaviors are then formed by $\pi_{b}=\alpha\pi_{e}+(1-\alpha)\pi^{+}$ with $\alpha\in{\{0.2, 0.4\}}$.
The true expected return of the target policy is approximated by a set of $2\times 10^{6}$ on-policy Monte-Carlo trajectories, truncated at $H=500$ to assure that the majority of the trials have stopped at the absorbing state before time step $H$.


\subsection{Results}

Reported here are the experiment results for MWLA, MSWLA, an on-policy, IS, and a naive averaging baseline algorithm. The on-policy algorithm (referred to as On-Policy below) estimates the expected return by the direct average over a set of trajectories generated by the target policy itself, and the naive average baseline algorithm (referred to as Naive Average below) does it by a direct average over a different set of trajectories generated by the behavior policy, all truncated at $H$. We also show the results of MWL applied to another set of simulated data.

\begin{figure}[H] 
\centering 
\includegraphics[width=0.9\textwidth]{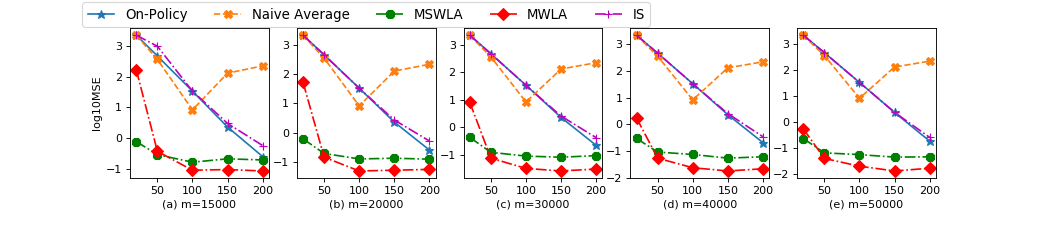}
\\{\footnotesize {\bf Agend}: The horizontal axis indicates the truncation levels $H$ and the vertical the logarithm \\of the MSEs.}
\includegraphics[width=0.9\textwidth]{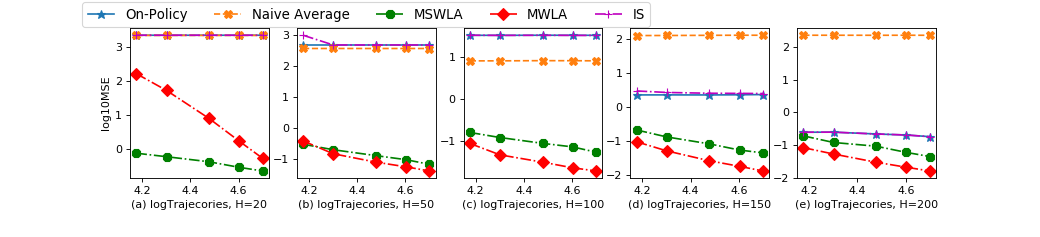}
\\{\footnotesize {\bf Agend}: The horizontal axis indicates the number of trajectories and the vertical the MSE, \\both are scaled in logarithm.}
\caption{MSE of the five algorithms ($\alpha=0.2$).\label{Fig.main1}}
\end{figure}

\begin{figure}[H] 
\centering 
\includegraphics[width=0.9\textwidth]{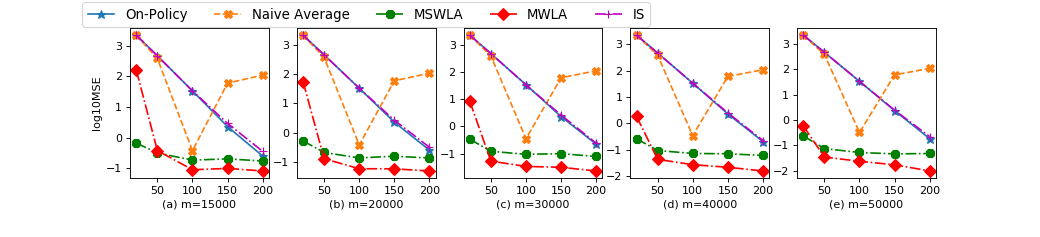}\\{\footnotesize {\bf Agend}: The horizontal axis indicates the truncation level $H$ and the vertical the logarithm\\ of the MSEs.}
\includegraphics[width=0.9\textwidth]{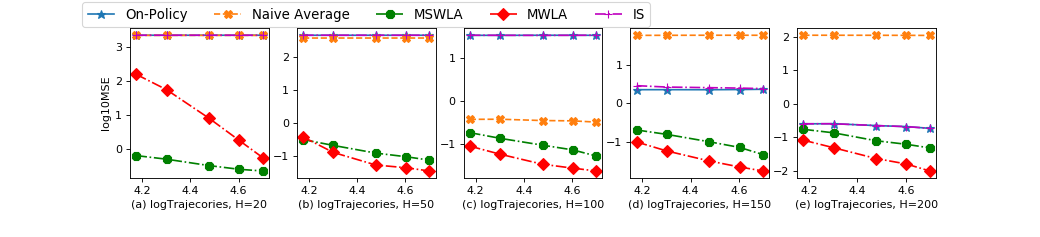}\\{\footnotesize {\bf Agend}: The horizontal axis indicates the number of trajectories and the vertical the MSE,\\ both are scaled in logarithm.}
\caption{MSEs of the five algorithms ($\alpha=0.4$).} 
\label{Fig.main3} 
\end{figure}

The first experiment is on the MSEs of the five methods under $m=15000$, $20000$, $30000$, $40000$ and $50000$ trajectories and a set of truncation levels $H=20,50,100,150,200$. A total of $100$ duplicates for every parameter combination are generated with different random seeds.
The results are visualized in Figures \ref{Fig.main1} (for $\alpha=0.2$) and \ref{Fig.main3} (for $\alpha=0.4$), where every graph corresponds to a single episode size $m$ in the upper panels, and a single truncation level $H$ in the lower panels. The MSEs of MWLA and MSWLA all decrease at the beginning and then vary slowly when the truncation level increases. MWLA is better than MSWLA to a moderate degree, and both are significantly lower than the on-policy, IS, and naive averaging baseline algorithms.

\begin{figure}[H] 
\centering 
\includegraphics[width=0.8\textwidth]{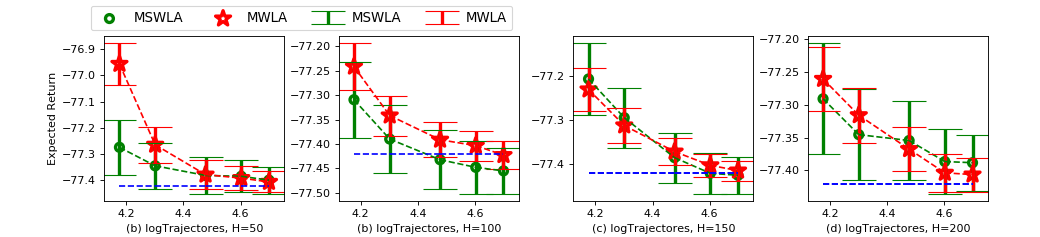}\\
{\footnotesize $\alpha=0.2$}\\
\includegraphics[width=0.8\textwidth]{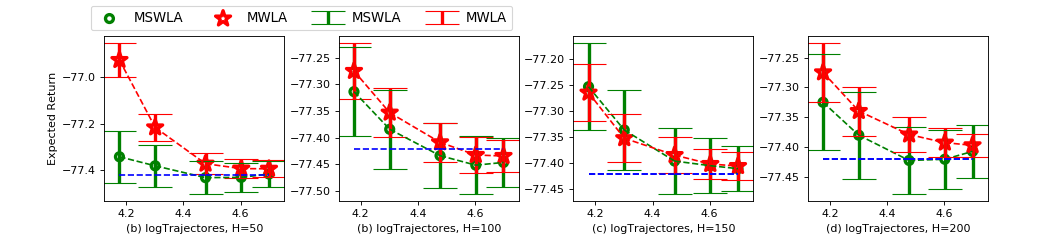} 
\\{\footnotesize$\alpha=0.4$}
\\{\footnotesize {\bf Agend}: 1. The horizontal axis indicates the number of trajectories and the vertical the MSE,\\ both are scaled in logarithm. 2. The blue lines represent the true expected returns.}
\caption{Estimated expected returns. } 
\label{Fig.main4} 
\end{figure}

The averaged estimated returns are also provided, see Figure \ref{Fig.main4}, together with {twice the standard errors of the estimates $\frac{1}{N}\sum_{i=1}^{N}\hat{R}^{i}_m\pm\frac{2S}{\sqrt{N}}$, corresponding to the 95\% confidence intervals, where N=100 is the number of duplicates, $\hat{R}^{i}_m$ is the estimated return of the $i$-th duplicate, and $S$ is the sample standard deviation of $\hat{R}^{i}_m, i=1,\dots,N$.} The estimates by  MSWLA and MWLA both approach the expected returns  as the numbers of trajectories get large. MSWLA has slightly smaller biases than MWLA but significantly larger fluctuations, giving rise to a higher MSE, as indicated by Figures \ref{Fig.main1} and \ref{Fig.main3}, even in the final graph in the bottom panel, with a quite small deviation of the averaged estimated returns from the expected return for the largest data size  due to the randomness of the data.

\begin{figure}[H] 
\centering 
\includegraphics[width=0.9\textwidth]{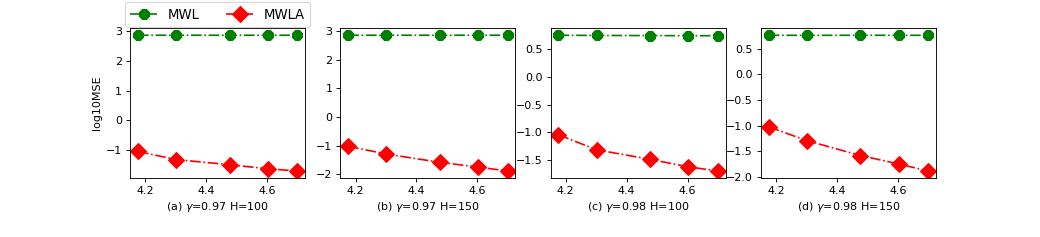}\\\includegraphics[width=0.9\textwidth]{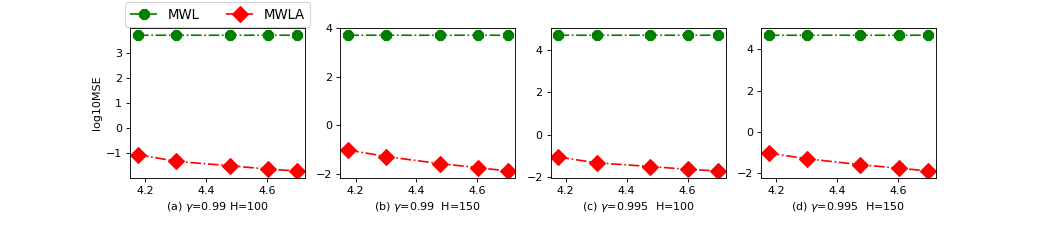} 
\\{\footnotesize {\bf Agend}: The horizontal axis indicates the number of trajectories and the vertical the MSE,\\ both are scaled in logarithm.}
\caption{The MSEs of MWL and MWLA algorithms ($\alpha=0.2$).} 
\label{Fig.main5} 
\end{figure}

The final experiment is to examine the performance of the MWL by \cite{uehara2020minimax} algorithm to estimate the expected undiscounted returns of absorbing MDPs by treating them as a special case of infinite-horizon MDPs with a subjectively designed large discount factor $\gamma$. The experiment is carried out under the policy mixing ratio $\alpha=0.2$, truncation levels $H =100$ and $150$, and numbers of trajectories $m = 15000$, $20000$, $30000$, $40000$, $50000$. We also produce $N=100$ duplicates for every parameter combination to empirically evaluate the MSEs. Two methods are employed to estimate the expected returns. One is the direct MWLA with trajectory data as above. The other is the MWL algorithm under large discount factors $\gamma=0.97$, $0.98$, $0.99$ and $0.995$, where the data consisting of all the tuples $(s,a,r,s')$ obtained by breaking the $m$ trajectories and the MSEs are computed using the errors between the estimates and the true expected (undiscounted) return. Here, we need to note that the MWL algorithm really estimates some quantity $A(\gamma,\pi_e)$ that is a function of the artificially associated discount factor $\gamma$ and the target policy $\pi_e$. The error $|A(\gamma,\pi_e)-R_{\pi_e}|$ between $A(\gamma,\pi_e)$ and the true value $R_{\pi_e}$ thus depends on the policy $\pi_e$ and, more importantly, the discount fact $\gamma$ also, so that there could exist some optimal $\gamma_0$, the value of which is certainly unknown to the agent because so is the MDP model $M$, to minimize $|A(\gamma,\pi_e)-R_{\pi_e}|$ and thus the MSE.  The MSE result is empirically depicted in Figure \ref{Fig.main5}, where the horizontal axis is again indicated by the logarithm of the trajectory numbers.
From the figure, we clearly have the following observations:
\begin{enumerate}[(1).]
\item The MSEs are almost constant for the number of trajectories we experimented with, quite possibly implying that, compared with the variance, bias contributes the most to an MSE.
 \item The MSEs of the MWL algorithm vary over different $\gamma$, meaning that different $\gamma$ gives rise to different errors of the MWL algorithm to the true value $R_{\pi_e}$ (caused mainly by the bias according to the previous item).

\item It appears that $\gamma_0=0.98$ is optimal in our experiment. However, it is unclear how to identify an optimal $\gamma_0$ theoretically, even approximately.
\item MWLA significantly performs better than MWL, which we attribute to the unbiasedness of the MWLA by taking $\gamma=1$.
\end{enumerate}

\section{Conclusions and discussions}\label{sec:6}

This paper addresses an OPE problem for absorbing MDPs. We propose the MWLA algorithm as an extension of the MWL algorithm by \cite{uehara2020minimax}. This MWLA algorithm proceeds with episodic data subject to truncations. The expected return of a target policy is estimated and an upper bound of its MSE is derived, in which the statistical error is related to both data size and truncation level. We also briefly discuss the MSWLA algorithm for situations where behavior policies are known. The numerical experiments demonstrate that the MWLA method has a lower MSE as the number of episodes and truncation length increase, significantly improving the accuracy of policy evaluation.

Conceptually, one can estimate the corresponding state-action value function $Q$ using estimated expected return, for example, by fitted-Q evaluation. The double robust (DR) estimation algorithm, which integrates learning weights and state-action value functions $Q$, is an effective and robust approach. It is now still unclear if a DR variant of the MWLA algorithm can be developed.

In statistics, confidence intervals are important to quantify the uncertainty of the point estimates. Recent work in the RL area includes \cite{shi2021deeply}, who propose a novel deeply-debiasing procedure to construct efficient, robust, and flexible confidence intervals for the values of target policies for infinite-horizon discounted MDPs and  \cite{dai2020coindice}, who develops a CoinDICE algorithm for calculating confidence intervals. It would be interesting to combine these methods with MWLA for absorbing MDPs.

Moreover, we would like to note that policy optimization is one of the crucial goals of RL. The policy optimization based on the MWL algorithms has been analyzed by \cite{liu2019off} that proposes an off-policy policy gradient optimization technique for infinite-horizon discounted MDP by using MSWL to correct state distribution shifts for the i.i.d. tuple data structure,  and \cite{huang2020convergence} that investigates the convergence of two off-policy policy gradient optimization algorithms based on state-action density ratio learning, among others. Therefore, it is important and possible to explore how the MWLA approach can be used in off-policy optimization for absorbing MDPs.

Finally, it should be noted that the state-action space considered in this paper is discrete.  This choice is based on the fact that the absorbing MDPs with discrete state-action space are prevalent in real-world applications. Moreover, our theoretical analysis heavily relies on the discrete feature of the state-action space as evidenced by e.g., the proof of Lemma \ref{Lemma 7} in Appendix. For cases involving continuous state-action spaces, it is quite common practice to employ approximations by e.g. linear or deep neural networks, so it would need additional efforts and considerations to extend the framework to continuous state-action spaces.

\bigskip

\noindent{\bf Acknowledgments}

The authors thank the editors and referees for their
helpful comments and suggestions.

\bigskip

\footnotesize

\bibliographystyle{plainnat}


\bibliography{refn}

\bigskip

\normalsize
\noindent{\large\bf Appendix}

\bigskip
\appendix

\section{Proof of Theorems in Section 3}\label{A1}

\noindent{\bf A.1. Proof of Theorem \ref{Theorem 1}.}

It suffices to prove the uniqueness. Note that, for all $(\bar s,\bar a)\in \mathcal{S}_0\times\mathcal{A}$,
\[
L(w,{\bf 1}_{(\bar s, \bar a)})=\mu(\bar s){\pi_e}(\bar a|\bar s)+\sum_{(s,a)\in\S_0\times\A}w( s, a)d_{{\pi_b}}( s, a)P(\bar s| s, a){\pi_e}(\bar a|\bar s)-w(\bar s, \bar a)d_{\pi_b}(\bar s, \bar a).
\]
By the condition that $d_{\pi_e}$ is the unique solution to \eqref{U2}, it follows that $ wd_{\pi_b}=d_{\pi_e} $, i.e. $ w= w_{\pi_e\over\pi_{b}}$.

\noindent{\bf A.2. Proof of Theorem \ref{Theorem 2}.}

With any policy $\pi$, associate a one-step forward operator $ \mathcal{L}_\pi $ on $\Xi$, defined by
 \beqlb\label{1step-forward}
 (\mathcal{L}_\pi q)(s,a)\doteq\sum_{(s',a')\in\S_0\times\A}P(s'|s,a)\pi(a'|s')q(s',a')=\E_\pi\left[q(s_{t+1},a_{t+1})|s_t=s,a_t=a\right]. \eeqlb
Then, ${\cal L}_\pi$ is linear over $\Xi$. Moreover, the state-action function with respect to $\pi$
 \beqlb\label{defQ2}
Q_{\pi}(s, a)&\doteq&\E_{(s,a),\pi}(\sum_{t=0}^{T-1}r(s_{t}, a_{t}))\nonumber\\
&=&R(s, a)+\sum_{s'\in S_0}P(s'|s,a)\mathrm{E}_{s', \pi}\left(\sum\limits_{t=1}^{T-1}r_{t} \right)\nonumber
\\&=&R(s, a)+\sum_{(s',a')\in \S_0\times \A}P(s'|s,a)\pi(a'|s')Q_{\pi}(s', a')\nonumber\\
&=&R(s, a)+(\mathcal{L}_\pi Q_\pi)(s, a)
\eeqlb
and, by definition \eqref{defL} of $L(w,q)$, the error function
 \beqlb\label{delayformL}
 L(w,q)=\sum_{s,a\in\mathcal{S}_0\times\mathcal{A}}w(s,a)\left[\mathcal{L}_{\pi_e} q(s,a)- q(s,a)\right] d_{\pi_b}(s,a)+\E_{s\sim \mu}\big[q(s,{\pi_e})\big].
 \eeqlb
By Theorem  \ref{Theorem 1}, we can further change the form of $L(w,q)$ as
\beqlb\label{delayformL-1}
 L(w,q)=\sum_{(s,a)\in\mathcal{S}_0\times\mathcal{A}}\left[w_{\pi_e/\pi_b}(s,a)-w(s,a)\right]\left[q(s,a)- (\mathcal{L}_{\pi_e} q)(s,a)\right] d_{\pi_b}(s,a).
 \eeqlb

By the definition, $V_{s',a',\pi_e}$ is the Q-function of the reward ${\bm 1}_{(s',a')}(s,a)$ under the policy $\pi_e$. Therefore, Equation \eqref{defQ2} states that
$$
V_{s',a',\pi_e}(s,a)-\mathcal{L}_{\pi_e}V_{s',a',\pi_e}(s,a)={\bm 1}_{(s',a')}(s,a).
$$
It then follows from putting the above into \eqref{delayformL-1} that, for every $(s',a')\in \S_0\times\A$,
\beqnn
    L(w,V_{s',a',\pi_e}) &=&\sum_{(s,a)\in\S_0\times\A} \left[w_{\pi_e\over\pi_{b}}(s,a)-w(s,a)\right]{\bm 1}_{(s',a')}(s,a)d_{\pi_{b}}(s,a)\\
    &=&d_{\pi_e}(s',a')-w(s',a')d_{\pi_{b}}(s',a').
\eeqnn
This gives the first equality of Equation (\ref{specialError}). The second can be examined similarly by considering the function \(V_{s',a',\pi_e}/d_{\pi_{b}}(s',a')\) to instead \(V_{s',a',\pi_e}\).

 We prove the remainder results in what follows.

(1). As a result of the first equality of part (1),
\[
|d_{\pi_e}(s',a')-w(s',a')d_{\pi_{b}}(s',a')|=|L(w,V_{s',a',\pi_e})|,
\]so that
\[
\|d_{\pi_e}-wd_{\pi_{b}}\|_\infty=\max_{(s',a')\in\S_0\times\A}|L(w,V_{s',a',\pi_e})|\leq \max_{q\in{\cal Q}}|L(w,q)|,
\] provided that $\{\pm V_{s',a',\pi_e}: (s',a')\in\mathcal{S}_0\times\mathcal{A}\}\subseteq \mathcal{Q}$,
The desired result (2) is thus proved.

(3). If $\{\pm V_{s',a',\pi_e}/d_{\pi_{b}}(s',a'): (s',a')\in\mathcal{S}_0\times\mathcal{A}\}\subseteq \mathcal{Q},
$ then, substitute $w^*$ for $w$ in the second equality of part (1) gives ruse to
$$
\|w_{\pi_e\over\pi_b}-w^*\|_\infty=\max_{(s',a')\in\S_0\times\A}|L(w^*,V_{s',a',\pi_e}/d_{\pi_{b}}(s',a'))|\leq  \max_{q\in\mathcal{Q}}|L(w^*, q)|=\min_{w\in\mathcal{W}}\max_{q\in\mathcal{Q}}|L(w,q)|.
$$

The proof is thus complete.

\bigskip

\section{Proof of Theorems in Section 4}
%
To begin with, the following two results are quoted for easy reference later on.
Denote by
$
\mathcal{F}\left(\left\{Z_{i}\right\}\right)=\left\{\left(f\left(Z_{1}\right), \ldots, f\left(Z_{n}\right)\right) \mid f \in \mathcal{F}\right\} \subset \mathbb{R}^{n}
$ a set generated by a function class $\mathcal{F}$ and a data set $\left\{Z_{1}, \ldots, Z_{n}\right\}$.
Then the empirical $\ell_{1}$-covering number $\mathcal{N}\left(\epsilon ; \mathcal{F}, \{Z_{i}\}_{i=1}^{n}\right)$ refers to the smallest number of balls of radius $\epsilon$ required to cover $\mathcal{F}\left(\left\{Z_{i}\right\}\right)$, where, the distance is computed by the empirical $\ell_{1}$-norm
$
\|f-g\|_{\mathcal{F}\left(\left\{Z_{i}\right\}\right)}:=\frac{1}{n} \sum_{i=1}^{n}\left|f\left(Z_{i}\right)-g\left(Z_{i}\right)\right| .
$

\begin{lemma} \label{Lemma 12.}(\citealp{pollard2012convergence}) Let $\mathcal{F}$ be a permissible class of  functions $\mathcal{Z} \rightarrow[-M, M]$ and $\left\{Z_{i}\right\}_{i=1}^{n}$ are i.i.d. samples from some distribution $P$. Then, for any given $\epsilon>0$,
\begin{equation}
\mathbb{P}\big(\sup _{f \in \mathcal{F}}\big|\frac{1}{n}\sum_{i=1}^{n} f\big(Z_{i}\big)-\mathrm{E}f(Z)\big|>\epsilon\big) \leq 8 \mathbb{E}\big[\mathcal{N}\left(\epsilon ; \mathcal{F},\{Z_i\}_{i=1}^n\right)\big] \exp \big(\frac{-n \epsilon^{2}}{512 M^{2}}\big).
\end{equation}
\end{lemma}
For a class of functions $\mathcal{F}$ on a measurable space equiped with a probability measure $\vartheta$, the covering numbers $\mathcal{N}(\varepsilon, \mathcal{F}, L^r(\vartheta))$ refers to the smallest number of balls of radius $\epsilon$ required to cover $\mathcal{F}$, where distances are measured in terms of the $L^r$-norm $\|f\|_{L^r(\vartheta)}:=(\int |f|^r d \vartheta)^{1/r}$ for all $f\in\mathcal{F}$.
The covering number can then be bounded in terms of the function class's pseudo-dimension:

\begin{lemma}\label{Lemma 13} (\citealp{sen2018gentle})
Suppose that $\mathcal{F}$ is a class of functions  with measurable envelope $F$ (i.e. $|f|\leq F$ for any $f\in\mathcal{F}$) and has a finite pseudo-dimension $D(\mathcal{F})$. Then, for any $r \geq 1$, and any probability measure $\vartheta$ such that $\|F\|_{L^r(\vartheta)}>0$,
$$\mathcal{N}(\varepsilon \|F\|_{L^r(\vartheta)}, \mathcal{F},L^r(\vartheta)) < K D(\mathcal{F}) (4e)^{D(\mathcal{F})}(2/\varepsilon)^{rD(\mathcal{F})},$$
where  $K$  is a universal constant and $0 < \varepsilon < 1$.
\end{lemma}

\medskip

Recall that in Theorem \ref{Theorem 3}, we address that the functional classes $\mathcal{W}$ and $\mathcal{Q}$ have finite pseudo-dimensions $D_{\mathcal{W}}$ and $D_{\mathcal{Q}}$,  $Q_{\pi_e}\in\mathcal{Q}$, and there exists a function $G$ satisfying that $|w|\leq G$, $|q|\leq G$ for all $w\in\mathcal{W}$, $q\in\mathcal{Q}$ and
          $\Lambda_1=\sum_{(s, a)\in\S_0\times \A} G(s, a)<+\infty,\;\; \Lambda_2:=\sum_{(s, a)\in\S_0\times \A} G^2(s, a)<+\infty$. In addition, we assume that there exists $\lambda_0>0$ such that $\mathrm{E}_{\mu,\pi_b}(e^{\lambda_0 T})=M_0<\infty$. Besides them, we also remind here that our data consist of i.i.d samples from $ M=(\mathcal{S}, \mathcal{A}, \mathcal{R}, P, \mu)$ under the behavior policy $\pi_b$. The proof of Theorem \ref{Theorem 3} proceeds in the following 3 parts: decomposition, evaluation, and optimization.

\medskip

\noindent{\bf B.1. Decomposition}

First decompose the estimate $\hat{R}_{\pi_e,m}$ defined in equation \eqref{EstimatedReturn} as
\beqlb\label{DcR}
   \hat{R}_{\pi_e,m}-R_{\pi_e}=I_1+I_2+I_3,
\eeqlb
where
 \beqnn
 I_1&=&\sum_{(s,a)\in\S_0\times\A}[\hat{w}_{m}(s,a)-w_{\pi_e\over\pi_b}(s,a)]R(s, a)d_{\pi_b}(s,a),
\\ I_2&=&\frac{1}{m}\sum\limits_{i=1}^{m}\sum_{(s,a)\in\S_0\times\A}\hat{w}_m(s,a)R(s, a)\left[\hat{d}^{i}_{\pi_{b}}(s,a)-d_{\pi_b}(s,a)\right],
 \eeqnn
 and
 \beqnn
 I_3=\frac{1}{m}\sum\limits_{i=1}^{m}\sum_{(s,a)\in\S_0\times\A}\hat{w}_m(s,a)\Big(\hat{R}^i(s,a)-R(s, a)\Big)\hat{d}^{i}_{\pi_{b}}(s,a).
 \eeqnn
By (\ref{defQ2}) and (\ref{delayformL-1}),
 $$I_1=\sum_{(s,a)\in\S_0\times\A}[\hat{w}_{m}(s,a)-w_{\pi_e\over\pi_b}(s,a)][Q_{\pi_e}(s,a)-(\mathcal{L}_{\pi_e} Q_{\pi_e})(s,a)]d_{\pi_b}(s,a)=-L(\hat{w}_m, Q_{\pi_e}).$$
Because $Q_{\pi_e}\in \mathcal{Q}$, it follows that
\beqnn
I_1^2&=&\big(L(\hat{w}_m,Q_{\pi_e})-\hat{L}_{m}(\hat{w}_{m},Q_{\pi_e} )+\hat{L}_{m}(\hat{w}_{m},Q_{\pi_e})\big)^{2}\\
  &\leq &2\big(L(\hat{w}_{m},Q_{\pi_e})-\hat{L}_{m}( \hat{w}_{m},Q_{\pi_e} )\big)^2+2\max_{q\in\mathcal{Q}}\hat{L}_{m}^2( \hat{w}_m,q),
\eeqnn
where $\hat{L}_m$ is defined in \eqref{f2} and we have used the fact  that
$$ \hat{L}_{m}^2( \hat{w}_{m},Q_{\pi_e})\leq \max_{q\in\mathcal{Q}}\hat{L}_{m}^2( \hat{w}_{m},q )= \min_{w\in\mathcal{W}}\max_{q\in\mathcal{Q}}\hat{L}_{m}^2( w, q).$$
Using the fact
$$\max_{q\in\mathcal{Q}}\hat{L}_{m}^2( \hat{w}_m,q)\leq 2\max_{q\in\mathcal{Q}}L^2( \hat{w}_m,q)+2\max_{q\in\mathcal{Q}}\left(\hat{L}_{m}( \hat{w}_m,q)-L( \hat{w}_m,q)\right)^2,$$
we further have
  \beqlb\label{Dci1}
   I_1^2&\leq& 6I_{11}^2 +4\min\limits_{w\in\mathcal{W}}\max\limits_{q\in\mathcal{Q}}L(w,q)^{2},
\eeqlb
where
 $$I_{11}^{2}=\sup\limits_{w\in\mathcal{W},q\in\mathcal{Q}}|L(w,q)-\hat{L}_{m}(w,q)|^{2}.$$
Substituting (\ref{defUb}) into the expression of $\hat{L}_m(w, q)$ in  \eqref{f2}, it follows that
\beqnn
\hat{L}_m (w,q)-\E_\mu(q(s, \pi_e ))
&=&\frac{1}{m}\sum_{i=1}^m \sum_{t=0}^{l_i -1}w(s_t^i, a_t^i)(q(s_{t+1}^i,\pi_e )-q(s_{t}^i,a^i_t)),
\eeqnn
{where  $q(\xi, \pi_e)=0$ is assumed}. Similarly,
\beqnn
L(w,q)-\E_\mu(q(s, \pi_e))=\E_{\mu,\pi_b}\Big(\sum_{t=0}^{T-1}w(s_t, a_t)(q(s_{t+1},\pi_e )-q(s_t,a_t))\Big).
\eeqnn
Define
$$\tilde{L}(w,q)=\E_{\mu,\pi_b}\Big(\sum_{t=0}^{T\wedge {H_\beta}-1}w(s_t, a_t)(q(s_{t+1},\pi_e )-q(s_t,a_t))\Big)$$
and
 $$\tilde{L}_m(w, q)=\frac{1}{m}\sum_{i=1}^m\sum_{t=0}^{T_i\wedge H_\beta-1}w(s_t^i, a_t^i)\big(q(s_{t+1}^i,\pi_e )-q(s_{t}^i,a^i_t)\big),$$
where $H_\beta$ is a constant specified later in B.2. Then
 \beqlb\label{DL}
 |\hat{L}_m(w,q)-L(w,q)|&=&|\hat{L}_m(w,q)-\E_\mu(q(s,\pi_e ))-(L(w,q)-\E_\mu(q(s,\pi_e )))|\nonumber
  \\&\leq|&\hat{L}_m(w,q)-\E_\mu(q(s,\pi_e ))-\tilde{L}_m(w,q)|+|\tilde{L}_m(w,q)-\tilde{L}(w,q)|\nonumber
  \\&&+|\tilde{L}(w,q)-(L(w,q)-\E_\mu(q(s,\pi_e )))|.
 \eeqlb

\medskip

\medskip

\noindent{\bf B.2. Evaluation}

\medskip

For any $\beta\geq M_0 e^{-\lambda_0 H}$, let $H_\beta=\min\{k:M_0e^{-\lambda_0k}\leq \beta\}=\lceil\ln(M_0/\beta)/\lambda_0\rceil$, where $\lceil x \rceil$ is the minimum integer no less than $x$. Obviously, $H_{\beta}\leq H$ and $M_0 e^{-\lambda_0 H_{\beta}}\leq \beta.$

\medskip

\noindent{\bf B.2.1. The upper bound of $\E(I_{11}^2)$}

An upper bound of $\E(I_{11}^2)$ is derived via a sequence of auxiliary results.


\begin{lemma}\label{Lemma 6} With the constant $\Lambda_1$, for any $\beta\geq M_0 e^{-\lambda_0 H}$,
\beqnn
\E(\sup_{w\in\mathcal{W}, q\in\mathcal{Q}}|\hat{L}_m(w, q)-\E_\mu(q(s,\pi_e ))-\tilde{L}_m(w, q)|^2)
 \leq\frac{4\Lambda_1^4 }{\lambda_0^2}\Big[\beta^2+\frac{2\beta}{m}\Big].
 \eeqnn
 \end{lemma}
\noindent{\bf Proof.} For any $w\in\mathcal{W}$ and $q\in\mathcal{Q}$,
 \beqnn
|\hat{L}_m(w, q)-\E_\mu(q(s,\pi_e ))-\tilde{L}_m(w, q)|&=&\Big|\frac{1}{m}\sum_{i=1}^m\sum_{t=T_i\wedge H_\beta}^{l_i -1}w(s_t^i, a_t^i)\left(q(s_{t+1}^i,\pi_e )-q(s_{t}^i,a^i_t)\right){\bm 1}_{\{T_i>H_\beta\}}\Big|.
 \eeqnn
 Recall the notation $l_i=T_i\wedge H$. Due to the assumption (1) in Theorem \ref{Theorem 3}, $w$ and $q$ are bounded by $\Lambda_1$, respectively,
 \beqnn
 |\hat{L}_m(w, q)-\E_\mu(q(s,\pi_e ))-\tilde{L}_m(w, q)|\leq\frac{2\Lambda_1^2}{m}\sum_{i=1}^m(T_i-H_\beta){\bm 1}_{\{T_i>H_\beta\}}.
 \eeqnn
Therefore,
\beqlb\label{lem6-1}
&&\E(\sup_{w\in\mathcal{W},q\in\mathcal{Q}}|\hat{L}_m(w, q)-\E_\mu(q(s,\pi_e ))-\tilde{L}_m(w, q)|^2)\nonumber\\
&\leq &\frac{4\Lambda_1^4}{m^2}\left(\sum_{i,j=1,i\not=j}^m \E[(T_i-H_\beta)(T_j-H_\beta){\bm 1}_{\{T_i,T_j>H_\beta\}}]+\sum_{i=1}^m \E[(T_i-H_\beta)^2{\bm 1}_{\{T_i>H_\beta\}}]\right)\nonumber\\
&=&\frac{4\Lambda_1^4}{m^2}\left(m(m-1) \E^2[(T_1-H_\beta){\bm 1}_{\{T_1>H_\beta\}}]+m\E[(T_1-H_\beta)^2{\bm 1}_{\{T_1>H_\beta\}}]\right),
\eeqlb
where the equality follows from the i.i.d. property of trajectories.  By further the inequality $x\vee x^2/2\leq e^x$ for every $x>0$,
 \beqlb\label{lem6-2}
 \E[(T_1-H_\beta){\bm 1}_{\{T_1>H_\beta\}}]&\leq& \frac{1}{\lambda_0}\E\left(e^{\lambda_0(T_1-H_\beta)}{\bm 1}_{\{T_1>H_\beta\}}\right)\leq \frac{M_0 e^{-\lambda_0 H_\beta}}{\lambda_0}\leq \frac{ \beta}{\lambda_0},
 \eeqlb
and
 \beqlb\label{lem6-3}
 \E[(T_i-H_\beta)^2{\bm 1}_{\{T_i>H_\beta\}}]&\leq& \frac{2}{\lambda_0^2}\E(e^{\lambda_0(T_1-H_\beta)}{\bm 1}_{\{T_1>H_\beta\}})
 \leq \frac{2M_0 e^{-\lambda_0 H_\beta}}{\lambda_0^2}\leq \frac{2\beta}{\lambda_0^2},
 \eeqlb
 substituting (\ref{lem6-2}) and (\ref{lem6-3})  into (\ref{lem6-1}) leads to the desired result.\qed

\begin{lemma}\label{Lemma 7}
There exists a constant $C_1$ independent of $m, H$ such that for $m\geq 2$,
\beqnn
\E(\sup_{w\in\mathcal{W}, q\in\mathcal{Q}}|\tilde{L}_m(w, q)-\tilde{L}(w, q)|^2)\leq C_1H_\beta^2\frac{\ln m}{m}. 
\eeqnn
\end{lemma}
\noindent{\bf Proof.}
For a representative trajectory $Z$ of the form \eqref{behavior trajectories}, denote by
$$\tilde{w}_q(Z)=\sum\limits_{t=0}^{T\wedge H_\beta-1}w(s_{t},a_{t})(q(s_{t+1}, \pi_e )-q(s_t, a_t)).$$ We have that
 $$\tilde{L}_m(w, q)-\tilde{L}(w, q)=\frac{1}{m}\sum_{i=1}^m \tilde{w}_q(Z_i)-\E(\tilde{w}_q(Z)).$$
It is easy to see that $|\tilde{w}_q|\leq 2\Lambda_1^2H_\beta$. Denote by
 $\mathcal{H}=\{\tilde w_q(Z):
  w\in\mathcal{W}, q\in\mathcal{Q}\}.$
The distance in $\mathcal{H}$ can be bounded by
\beqnn
&&\frac{1}{m}\sum\limits_{i=1}^{m}\Big|
\sum_{t=0}^{T_{i}\wedge H_\beta-1}w_1(s_t^{i}, a_t^{i})(q_1(s_{t+1}^{i},\pi_e )-q_1(s_t^{i},a_t^{i}))
\\&&\qquad\qquad\qquad-\sum_{t=0}^{T_{i}\wedge H_\beta-1}w_2(s_t^{i}, a_t^{i})(q_2(s_{t+1}^{i},\pi_e )-q_2(s_t^{i},a_t^{i}))
\Big|
\\&&\quad\leq 2\Lambda_1H_\beta(||w_1-w_2||_\infty+\|q_1-q_2\|_\infty).
\eeqnn
Note that, in our setting,
\beqnn
||w_1-w_2||_\infty &\leq&\|w_1-w_2\|_2\leq \sum_{(s,a)\in\S_0\times\A}2G(s,a)|w_1(s,a)-w_2(s, a)|
\\&\leq& 2\Lambda_1\sum_{(s,a)\in\S_0\times\A}\frac{G(s,a)}{\Lambda_1}|w_1(s,a)-w_2(s, a)|,
\eeqnn
where at the second inequality, we use the assumption (\ref{Assumption4-2}). Let $\vartheta:=(\vartheta(s,a))$ be the probability on $\S_0\times\A$ such that $\vartheta(s, a):=G(s, a)/\Lambda_1$. We get that
  \beqnn
||w_1-w_2||_\infty \leq 2\Lambda_1\|w_1-w_2\|_{L^1(\vartheta)},
\eeqnn
The same arguments imply that
 \beqnn
||q_1-q_2||_\infty \leq 2\Lambda_1\|q_1-q_2\|_{L^1(\vartheta)}.
\eeqnn
As a result,
\beqnn
\mathcal{N}(4\Lambda_1^2H_\beta(\epsilon_1+\epsilon_2),\mathcal{H}, {\{Z_{i}\}}_{i=1}^{m})
&\leq& \mathcal{N}(\epsilon_1,\mathcal{W}, L^1(\vartheta))\mathcal{N}(\epsilon_2,\mathcal{Q}, L^1(\vartheta)).
\eeqnn
Note that $\|G\|_{L^1(\vartheta)}=\sum g^2(s,a)/\Lambda_1=\Lambda_2/\Lambda_1$. Due to the assumption (2) in Theorem \ref{Theorem 3}, from Lemma \ref{Lemma 13} we get that
\beqnn
\mathcal{N}_{1}(4H_\beta\Lambda_1^2(\epsilon_1+\epsilon_2),\mathcal{H}, {\{Z_{i}\}}_{i=1}^{m})
&\leq& K^2D_{\mathcal{W}}D_{\mathcal{Q}}\left(\frac{8e\Lambda_2}{\Lambda_1\epsilon_1}\right)^{D_{\mathcal{W}}}\left(\frac{8e \Lambda_2}{\Lambda_1\epsilon_2}\right)^{D_{\mathcal{Q}}}.
\eeqnn
Taking $\epsilon_1=\epsilon_2={\epsilon\over 64 H_\beta\Lambda_1^2}$, we have
$$\mathcal{N}_{1}\left(\frac{\epsilon}{8},\mathcal{H}, {\{Z_{i}\}}_{i=1}^{m}\right)\leq\frac{{M}}{\epsilon^{D_\mathcal{W}+D_{\mathcal{Q}}}},$$
where ${M}=KD_{\mathcal{W}}D_{\mathcal{Q}}(512e H_\beta\Lambda_1\Lambda_2)^{D_\mathcal{W}+D_\mathcal{Q}}.$
By Pollard's tail inequality (Lemma \ref{Lemma 12.}),
\beqlb\label{ieql2}
    &&P\left(\sup\limits_{w\in \mathcal W,q\in\mathcal {Q}} \left|\frac{1}{m}\sum\limits_{i=1}^{m}h_{w,q}(Z_{i})-\mathrm{E} h_{w,q}(Z)\right|> \epsilon\right)\leq \frac{8{M}}{\epsilon^{D_\mathcal{W}+D_\mathcal{Q}}}{\exp\big(\frac{-m\epsilon^{2}}{2048\Lambda_1^4H_\beta^2}\big)}.\qquad
\eeqlb
For any $m>1$, let $x_0=\frac{(32\Lambda_1^2H_\beta)^2(D_\mathcal{W}+D_\mathcal{Q})\ln m}{m}$. Then, by (\ref{ieql2}), we have that
 \beqnn
&&\E(\sup_{w\in\mathcal{W}, q\in\mathcal{Q}}|\tilde{L}_m(w, q)-\tilde{L}(w, q)|^2)\\
  &\leq & \int_0^\infty\P(\sup_{w\in\mathcal{W}, q\in\mathcal{Q}}|\tilde{L}_m(w, q)-\tilde{L}(w, q)|\geq \sqrt x)d x.
 \\&\leq &\int_0^{x_0}d x+\int_{x_0}^\infty  \frac{8{M}}{x^{(D_\mathcal{W}+D_\mathcal{Q})/2}}\exp\big(\frac{-m x}{2048(\Lambda_1^2H_\beta)^2}\big)d x
 \\&\leq& x_0+\frac{8{M}}{x_0^{(D_\mathcal{W}+D_\mathcal{Q})/2}}\int_{x_0}^\infty  \exp\big(\frac{-m x}{2048(\Lambda_1^2H_\beta)^2}\big)d x
 \\&=&\frac{(32 \Lambda_1^2H_\beta)^2}{m}\Big[(D_\mathcal{W}+D_\mathcal{Q})\ln m+\frac{16M}{\left[(32\Lambda_1^2H_\beta)^2\ln m(D_\mathcal{W}+D_\mathcal{Q})\right]^{(D_\mathcal{W}+D_\mathcal{Q})/2}}\Big],
 \eeqnn
which implies the desired result.\qed

\begin{lemma}\label{Lemma 8} 
There exists a constant $C_3$ independent of $m, H$ such that for $m\geq 2$,
\beqnn
\E(\sup_{w\in\mathcal{W}, q\in\mathcal{Q}}|\hat{L}_m(w, q)-L(w, q)|^2)\leq C_3\Big(\beta^2+(1-\ln\beta+\ln^2\beta)\frac{\ln m}{m}+\frac{\beta}{m}\Big). 
\eeqnn
\end{lemma}
\noindent{\bf Proof}. By (\ref{DL}),
 \beqlb\label{lem9}
\E[\sup_{w\in\mathcal{W},q\in\mathcal{Q}}|\hat{L}_m(w, q)-L(w, q)|^2]&\leq& 3\E(\sup_{w\in\mathcal{W},q\in\mathcal{Q}}|\hat{L}_m(w,q)-\E_\mu(q(s,\pi_e ))-\tilde{L}_m(w,q)|^2)\nonumber
\\&&+3\E(\sup_{w\in\mathcal{W},q\in\mathcal{Q}}|\tilde{L}(w, q)-(L(w, q)-\E_\mu(q(s,\pi_e )))|^2)\nonumber\\
&&+3\sup_{w\in\mathcal{W},q\in\mathcal{Q}}|\tilde{L}_{m}(w, q)-\tilde{L}(w, q)|^2.
 \eeqlb
Note that for any $w\in\mathcal{W}$,
 \beqlb\label{p3}
 |\tilde{L}(w, q)-(L(w, q)-\E_\mu(q(s,\pi_e )))|&=&\E\Big(\sum_{T\wedge H_\beta}^{T-1}w(s_t,a_t)(q(s_{t+1}, \pi_e )-q(s_t, a_t))\Big)\nonumber
 \\&\leq &2\Lambda_1^2\E((T-H_\beta){\bm 1}_{\{T\geq H_\beta\}})\nonumber\\
& \leq & 2\Lambda_1^2\E\left(\frac{e^{\lambda_0(T-H_\beta)}}{\lambda_0}{\bm 1}_{\{T\geq H_\beta\}}\right)\nonumber\\
&\leq& \frac{2\Lambda_1^2}{\lambda_0}\beta,
 \eeqlb
where the last inequality comes from the assumption (5) in Theorem \ref{Theorem 3} and the setting of $H_\beta$. Substituting (\ref{p3}) into (\ref{lem9}) and applying Lemma \ref{Lemma 6} and Lemma \ref{Lemma 7}, we get that
 \beqlb\label{lem9-1}
  \E(\sup_{w\in\mathcal{W},q\in\mathcal{Q}}|\hat{L}_m(w, q)-L(w, q)|^2)&\leq& 3M_3^2\mu^2 +3 \left(M_3^2\beta^2+\frac{2M_3^2}{m}\beta\right)+3C_1 H_\beta^2\frac{\ln m}{m},\nonumber
    \eeqlb
where $M_3=2\Lambda_1^2/\lambda_0$.  Since $H_\beta\leq\frac{\ln (M_0/\beta)}{\lambda_0}+1$, it follows from (\ref{lem9}) that
  \beqnn
  &&\E(\sup_{w\in\mathcal{W},q\in\mathcal{Q}}|\hat{L}_m(w, q)-L(w, q)|^2)
  \\&\leq&\frac{3C_1}{\lambda_0^2}\left((\ln\beta)^2-2(\ln M_0+\lambda_0)\ln\beta+(\ln M_0+\lambda_0)^2\right)\frac{\ln m}{m}+6M_3^2\beta^2 +\frac{6M_3^2}{m}\beta
  \\&\leq&\frac{3C_1}{\lambda_0^2}\left((\ln\beta)^2-2C_2\ln\beta+C_2^2\right)\frac{\ln m}{m}+6M_3^2\beta^2 +\frac{6M_3^2\beta}{m},
   \eeqnn
 where $C_2=\ln M_0+\lambda_0$.  Let
 $$C_3=\max\left\{\frac{3C_1}{\lambda_0^2},\frac{3C_1}{\lambda_0^2}C_2, \frac{3C_1}{\lambda_0^2}C_2^2, 6M_3^2\right\},$$
 we can readily get the desired result.\qed

\medskip

\noindent{\bf B.2.2. The upper bounds of $\E(I_2^2)$ and $\E(I_3^2)$}

\begin{lemma}\label{Lemma 9} Denote $\sup|R(s, a)|$ by $R_{\max}$. We have that $$\E(I_2^2)\leq\frac{2R_{\max}^2K_0^2M_0}{\lambda_0^2}(\frac{1}{m}+M_0e^{-2\lambda_0 H}).$$
\end{lemma}
\begin{proof}
Define a truncated occupancy measure $\tilde{d}_{\pi_b}(s, a)=\E_{\mu,\pi_b}(\sum_{t=0}^{T\wedge H-1}{\bm 1}_{(s, a)}(s_t, a_t)).$ Then
\beqnn
 I_2&=&\frac{1}{m}\sum\limits_{i=1}^{m}\sum_{(s,a)\in\S_0\times\A}\hat{w}_m(s,a)R(s, a)[\hat{d}^{i}_{\pi_{b}}(s,a)-\tilde{d}_{\pi_b}(s,a)]
 \\&&+\sum_{(s,a)\in\S_0\times\A}\hat{w}_{m}(s,a)R(s, a)[\tilde{d}_{\pi_b}(s,a)-{d}_{\pi_b}(s,a)]
\eeqnn
and hence
 \beqnn
 I_2^2&\leq& 2\Big[\sum_{(s,a)\in\S_0\times\A}\hat{w}_m(s,a)R(s, a)\Big(\frac{1}{m}\sum\limits_{i=1}^{m}\hat{d}^{i}_{\pi_{b}}(s,a)-\tilde{d}_{\pi_b}(s,a)\Big)\Big]^2
 \\&&+2\Big[\sum_{(s,a)\in\S_0\times\A}\hat{w}_m(s,a)R(s, a)\Big(\tilde{d}_{\pi_b}(s,a)-d_{\pi_b}(s, a)\Big)\Big]^2\\
 &\leq& 2\sum_{(s,a)\in\S_0\times\A}\hat{w}_m(s,a)^2R(s, a)^2\sum_{(s,a)\in\S_0\times\A}\Big(\frac{1}{m}\sum\limits_{i=1}^{m}\hat{d}^{i}_{\pi_{b}}(s,a)-\tilde{d}_{\pi_b}(s,a)\Big)^2
 \\&&+2\Big[\sum_{(s,a)\in\S_0\times\A}\hat{w}_m(s,a)R(s, a)\E_{\mu,\pi_b}(\sum_{t=T\wedge H}^{T-1}{\bm 1}_{(s. a)}(s_t, a_t))\Big]^2,
 \eeqnn
where the last inequality follows from H\"{o}lder's inequality. Invoking the bounds from $\hat w_m$ and $R$, we get that
  \beqnn
 \E(I_2^2)&\leq &2R_{\max}^2\Lambda_1^2\Big\{\sum_{(s,a)\in\S_0\times\A}\E\Big(\frac{1}{m}\sum\limits_{i=1}^{m}\hat{d}^{i}_{\pi_{b}}(s,a)-\tilde{d}_{\pi_b}(s,a)\Big)^2
\\&&\qquad\quad\qquad +\Big(\E_{\mu,\pi_b}\Big[\sum_{(s,a)\in\S_0\times\A}\sum_{t=T\wedge H}^{T-1}{\bm 1}_{(s,a)}(s_t, a_t)\Big]\Big)^2\Big\}
\\&\leq& 2R_{\max}^2\Lambda_1^2\Big\{\frac{1}{m}\sum_{(s,a)\in\S_0\times\A}{\rm Var}(\hat{d}^{i}_{\pi_{b}}(s,a)))+\Big(\E_{\mu,\pi_b}((T-H){\bm 1}_{\{T>H\}})\Big)^2\Big\},
\eeqnn
where the first inequality is due to $\hat{w}_m\in\mathcal{W}$ and the second inequality follows from the fact that $d_{\pi_b}^{i}(s, a), 1\leq i\leq m,$ are i.i.d and have  expectation $\tilde{d}_{\pi_b}(s, a)$. Observing that
\beqnn
{\rm Var}(\hat{d}^{i}_{\pi_{b}}(s,a)))&=&{\rm Var}_{\mu,\pi_b}(\sum_{t=0}^{T\wedge H-1}{\bm 1}_{(s,a)}(s_t, a_t))\leq \E_{\mu,\pi_b}\Big(\Big(\sum_{t=0}^{T\wedge H-1}{\bm 1}_{(s,a)}(s_t, a_t)\Big)^2\Big)
\\&\leq& \E_{\mu,\pi_b}\Big((T\wedge H)\sum_{t=0}^{T\wedge H-1}{\bm 1}_{(s,a)}(s_t, a_t)\Big),
\eeqnn
we obtain that
\beqnn
&&\sum_{(s,a)\in\S_0\times\A}{\rm Var}(\hat{d}^{i}_{\pi_{b}}(s,a)))\leq \E_{\mu,\pi_b}\Big((T\wedge H)\sum_{(s,a)\in\S_0\times\A}\sum_{t=0}^{T\wedge H-1}{\bm 1}_{(s,a)}(s_t, a_t)\Big)\leq \E_{\mu,\pi_b}\big(T^2\big).
 \eeqnn
Since $\E_{\mu,\pi_b}\big(T^2\big)\leq\frac{2M_0}{\lambda_0^2}$ and $\E_{\mu,\pi_b}((T-H){\bm 1}_{\{T>H\}})\leq \frac{M_0}{\lambda_0}e^{-\lambda_0 H}$, we arrive at the conclusion.
\end{proof}


\begin{lemma}\label{Lemma 10} We have that $$\E(I_3^2)\leq\frac{R_{\max}^2\Lambda_1^2M_0}{\lambda_0 m}.$$
\end{lemma}
\noindent{\bf Proof.} Noting that
\beqnn
I_3&=&\sum_{(s,a)\in\S_0\times\A}\hat{w}_m(s, a)\Big[\frac{1}{m}\sum_{i=1}^m (\hat{r}^i(s, a)-R(s, a))\hat{d}^i_{\pi_b}(s, a)\Big],
\eeqnn
applying the H\"{o}lder inequality, we have that
 \beqnn
I_3^2&\leq&\sum_{(s,a)\in\S_0\times\A}\hat{w}_m^2(s, a)\sum_{(s,a)\in\S_0\times \A}\Big[\frac{1}{m}\sum_{i=1}^m (\hat{r}^i(s, a)-R(s, a))\hat{d}^i_{\pi_b}(s, a)\Big]^2
\\&\leq& \Lambda_1^2\sum_{(s,a)\in\S_0\times \A}\Big[\frac{1}{m}\sum_{i=1}^m (\hat{r}^i(s, a)-R(s, a))\hat{d}^i_{\pi_b}(s, a)\Big]^2.
\eeqnn
Therefore,
 \beqnn
  \E [I_{3}^{2}]&\leq&
  \Lambda_1^2\E\left[\sum_{(s,a)\in\S_0\times \A}\E\left(\Big[\frac{1}{m}\sum_{i=1}^m (\hat{r}^i(s, a)-R(s, a))\hat{d}^i_{\pi_b}(s, a)\Big]^2\Big|\hat{d}^i_{\pi_b}(s, a),i=1,\cdots,m\right)\right].
 \eeqnn
Because $\hat{d}^i_{\pi_b}(s, a),i=1,\cdots,m$ are i.i.d and $r^i(s,a)$ follows  distribution $R(s, a)$ and is independent of $\hat{d}^i_{\pi_b}(s, a),i=1,\cdots,m$, it follows that
 \beqnn
 &&\E\left(\Big[\frac{1}{m}\sum_{i=1}^m (\hat{r}^i(s, a)-R(s, a))\hat{d}^i_{\pi_b}(s, a)\Big]^2\Big|\hat{d}^i_{\pi_b}(s, a),i=1,\cdots,m\right)
 \\&=&\frac{1}{m^2}\sum_{i=1}^m (\hat{d}^i_{\pi_b}(s, a))^2{\rm Var}\Big(\hat{r}^i(s, a)\Big|\hat{d}^i_{\pi_b}(s, a)\Big)
 \\&=&\frac{1}{m^2}\sum_{i=1}^m (\hat{d}^i_{\pi_b}(s, a))^2{\rm Var}\Big(\frac{\sum_{t=0}^{l_i -1} r^i_t{\bm 1}_{(s,a)}(s_t^i, a_t^i)}{\hat{d}^i_{\pi_b}(s, a)}\Big|\hat{d}^i_{\pi_b}(s, a)\Big).
 \eeqnn
When $\hat{d}^i_{\pi_b}(s, a)$ is given, $\sum_{t=0}^{l_i -1} r^i_t{\bm 1}_{(s,a)}(s_t^i, a_t^i)$ is the sum of $\hat{d}^i_{\pi_b}(s, a)$ random variables who are independent with the same distribution $R(s, a)$. Hence
 \beqnn
 {\rm Var}\Big(\hat{r}^i(s, a)\Big|\hat{d}^i_{\pi_b}(s, a)\Big)=\frac{{\rm Var}_{\mathcal{R}(s, a)}(r)}{\hat{d}^i_{\pi_b}(s, a)} \leq\frac{R^2_{max}}{\hat{d}^i_{\pi_b}(s, a)}.
 \eeqnn
Therefore
 \beqnn
\E(I_{3}^2)&\leq&\frac{R_{\max}^2\Lambda_1^2}{m^2}\E\Big(\sum_{(s, a)\in\S_0\times\A}\sum_{i=1}^m\hat{d}^i_{\pi_b}(s, a)\Big)= \frac{R_{\max}^2\Lambda_1^2}{m^2}\E\Big(\sum_{i=1}^m\sum_{(s, a)\in\S_0\times\A}\hat{d}^i_{\pi_b}(s, a)\Big)\nonumber
\\&=&\frac{R_{\max}^2\Lambda_1^2}{m^2}\E\Big(\sum_{i=1}^m (l_i -1)\Big)\leq \frac{R_{\max}^2\Lambda_1^2}{m}\E_{\mu,\pi_b}(T),
 \eeqnn
which implies the desired result, since $\E_{\mu,\pi_b}(T)\leq\frac{M_0}{\lambda_0}$.\qed
\medskip

\noindent{\bf B.3. Optimization}

\medskip

Based on the above discussion, we get that for any truncation level $H$ and $\beta$ such that $M_0e^{-\lambda_0 H}\leq \beta$,
\beqnn
\E(|\hat{R}_m-R_{\pi_e}|^2)&\leq &2\E(I_1^2)+4\E(I_2^2)+4\E(I_3^2)
\\&\leq& 8\min\limits_{w\in\mathcal{W}}\max\limits_{q\in\mathcal{Q}}L(w,q)^{2}+12\E(I_{11}^2)+4\E(I_2^2)+4\E(I_3^2)
\\&\leq & 8\min\limits_{w\in\mathcal{W}}\max\limits_{q\in\mathcal{Q}}L(w,q)^{2}+12C_3\Big(\beta^2+(1-\ln \beta+(\ln\beta)^2)\frac{\ln m}{m}+\frac{\beta}{m}\Big)
\\&&+ \frac{4R_{\max}^2\Lambda_1^2M_0}{m\lambda_0}+\frac{8\Lambda_1^2R_{\max}^2M_0}{\lambda_0^2}(\frac{1}{m}+M_0e^{-2\lambda_0H}),
\eeqnn
where the first inequality follows from (\ref{DcR}) and the simple inequality $(a+b)^2\leq 2a^2+2b^2$, the second inequality follows from (\ref{Dci1}) and the last inequality is due to Lemma \ref{Lemma 8}-Lemma \ref{Lemma 10}. Letting $$C_4=\max\{24C_3+\frac{4R_{\max}^2\Lambda_1^2M_0}{\lambda_0}+\frac{8\Lambda_1^2R_{\max}^2M_0}{\lambda_0^2}, \frac{8\Lambda_1^2R_{\max}^2M_0^2}{\lambda_0^2}\},$$
we have that, for any truncation level $H$ and $\beta\geq M_0e^{-\lambda_0 H}$,
 \beqlb\label{errR}
&&\E(|\hat{R}_{m}-R_{\pi_e}|^2)\leq 8\min\limits_{w\in\mathcal{W}}\max\limits_{q\in\mathcal{Q}}L(w,q)^{2}+C_4\big(G(\beta, m)+e^{-2\lambda_0H}\big),
\eeqlb
where
 $$G(x, m):=x^2+(1-\ln x+\ln^2x)\frac{\ln m}{m},$$
for $x\in(0, +\infty), m\geq 1$. Note that
 $$G'_x(x, m)=2x+\frac{\ln m(2\ln x-1)}{mx},\quad G''_{xx}(x,m)=2+\frac{\ln m(3-2\ln x)}{mx^2}>0,$$
which combining the fact $G'_x(1, m)>1$ and $\lim\limits_{x\to 0+}G'_x(x, m)=-\infty$ implies that there exists a unique $H_m\in(0, 1)$ such that
 $$2mH_m^2+2\ln m\ln H_m-\ln m=0,$$
which implies $G'_x(H_m, m)=0$ and hence for all $x\in(0, +\infty)$,
 $$G(H_m, m)\leq G(x, m).$$
Moreover, $G(x, m)$ is decreasing for $x\in(0, H_m)$ while increasing for $x\in(H_m, +\infty)$.

\begin{lemma}\label{Lemma 11} For any $m\geq e$, $\sqrt{\ln m/(2m)}\leq H_m\leq \sqrt{e \ln ^2m/m}$ and there exists constants $0<k_1<k_2$ such that
     $$k_1\frac{\ln^3 m}{m}\leq G(H_m, m)\leq k_2\frac{\ln^3 m}{m}.$$
     \end{lemma}
\noindent{\bf Proof.} From $G'_\mu(H_m, m)=0$, we know that $H_m$ is a solution of
 $$H_m(x):=2mx^2+2\ln m\ln x-\ln m=0.$$
 Note that $H_m(x)$ is increasing on $(0, 1]$ and
 \beqnn
 H_m\left(\sqrt{\frac{\ln m}{2m}}\right)&=&2\ln m\ln\left(\sqrt{\frac{\ln m}{2m}}\right)<0,
 \\ H_m\left(\sqrt{\frac{e\ln^2 m}{m}}\right)&=&(2e-1)\ln^2 m+2\ln m\ln\ln m>0.
  \eeqnn
We have that $\sqrt{\ln m/(2m)}\leq H_m\leq \sqrt{e \ln ^2m/m}$.

 To prove the second assertion, we note that
 \beqnn
 G(H_m, m)\leq G\left(\sqrt{\frac{\ln^3 m}{m}}, m\right)\leq \frac{\ln^3 m}{m}+(1+\frac{1}{2}\ln m+\frac{1}{4}\ln^2 m)\frac{\ln m}{m}\leq 3\frac{\ln^3 m}{m}.
 \eeqnn
On the other hand, when $x<\sqrt{\ln^3 m/m}$,
 \beqnn
 G(x, m)&\geq& (1-\ln x+\ln^2x)\frac{\ln m}{m}\geq\frac{1}{4}\left(\ln\frac{\ln^3 m}{m}\right)^2\frac{\ln m}{m}
 \\&\geq &\frac{\ln^3 m}{4m}\Big[1-3\frac{\ln\ln m}{\ln m}\Big]^{2}\geq \frac{(e^3-9)^{2}}{4e^6}\frac{\ln^3 m}{m},
 \eeqnn
for $m\geq e^{e^3}$. Noting that $\ln^3 m/m>0$ for all $m>e$, we can find a constant $k'$ such that
  $$G(x, m)\geq k'\frac{\ln^3 m}{m},$$
for all $x\in(0, \sqrt{\ln^3 m/m})$ and $m>e$. Moreover, when $x\geq \sqrt{\ln^3 m/m}$,
 \beqnn
 G(x, m)> x^2\geq \frac{\ln^3 m}{m}.
 \eeqnn
Consequently, there exists a constant $k_1$ such that $G(H_m, m)\geq k_1\ln^3 m/m.$\qed

Now we are at the position to finish the proof of Theorem \ref{Theorem 3}.

\noindent{\bf Proof.} From (\ref{errR}) and Lemma \ref{Lemma 11},  it follows that
if $H_m\geq M_0 e^{-\lambda_0 H}$, there is a constant $C$ independent of $m, H$ such that
 \beqnn
\E(|\hat{R}_m-R_{\pi_e}|^2)
&\leq&8\min_{w\in\mathcal{W}, q\in\mathcal{Q}}L^2(w, q)+ C_4\big(\min_{{\beta\geq M_0 e^{-\lambda_0 H}}}G(\beta, m)+e^{-2\lambda_0 H}\big)
\\&=&8\min_{w\in\mathcal{W}, q\in\mathcal{Q}}L^2(w, q)+ C_4(G(H_m, m)+e^{-2\lambda_0 H})
\\&\leq& 8\min_{w\in\mathcal{W}, q\in\mathcal{Q}}L^2(w, q)+ C \frac{\ln^3 m}{m},
\eeqnn
since $H_m\geq M_0 e^{-\lambda_0 H}$ implies that
 \beqnn
 e^{-2\lambda_0 H}\leq\frac{H_m^2}{M_0}\leq \frac{e}{M_0}\frac{\ln^2 m}{m}.
 \eeqnn
When $H_m<M_0 e^{-\lambda_0 H}$,
 \beqnn
&&\E(|\hat{R}_m-R_{\pi_e}|^2)\leq8\min_{w\in\mathcal{W}, q\in\mathcal{Q}}L^2(w, q)+ C_4(G(M_0e^{-\lambda_0 H}, m)+e^{-2\lambda_0 H})\nonumber
\\&&\quad=8\min_{w\in\mathcal{W}, q\in\mathcal{Q}}L^2(w, q)+ C_4\Big((1+M_0^2)e^{-2\lambda_0 H}+(1-\ln M_0+\lambda_0 H+(\ln M_0-\lambda_0H)^2)\frac{\ln m}{m}\Big)\nonumber
\\&&\quad\leq 8\min_{w\in\mathcal{W}, q\in\mathcal{Q}}L^2(w, q)+C\Big(e^{-2\lambda_0 H}+ H^2\frac{\ln m}{m}\Big),
\eeqnn
for some constant $C$ independent of $m, H$.\qed

From Theorem \ref{Theorem 3}, it is easy to get Therorem \ref{Corollary 4}. We briefly state the proof as follows.

\noindent{\bf Proof of Theorem \ref{Corollary 4}.} When $Q_{\pi_e}\not\in\mathcal{Q}$, (\ref{Dci1}) does not hold but can be adjusted as follows. Since
\beqnn
I_1^2&=&L^2(\hat{w}_m, Q_{\pi_e})=(L(\hat{w}_m, Q_{\pi_e}-q)+L(\hat{w}_m, q))^2\leq 2(L^2(\hat{w}_m, q)+L^2(\hat{w}_m, Q_{\pi_e}-q)),
\eeqnn
for any $q\in\mathcal{Q}$, we have that,
 \beqnn
I_1^2&\leq & 2\max_{q\in\mathcal{Q}}L^2(\hat{w}_m, q)+2\min_{q\in\mathcal{Q}}L^2(\hat{w}_m, Q_{\pi_e}-q).
\\&\leq & 2\max_{q\in\mathcal{Q}}(L(\hat{w}_m, q)-\hat{L}_m(\hat{w}_m, q)+\hat{L}_m(\hat{w}_m, q))^2+2\max_{w\in\mathcal{W}}\min_{q\in\mathcal{Q}}L^2(w, Q_{\pi_e}-q)
\\&\leq& 4\max_{q\in\mathcal{Q}}(L(\hat{w}_m, q)-\hat{L}_m(\hat{w}_m, q))^2+4\max_{q\in\mathcal{Q}}\hat{L}_m^2(w, q)+2\max_{w\in\mathcal{W}}\min_{q\in\mathcal{Q}}L^2(w, Q_{\pi_e}-q),
\eeqnn
for any $w\in\mathcal{W}$. Consequently,
 \beqnn
 I_1^2&\leq& 4\max_{w\in\mathcal{W}, q\in\mathcal{Q}}(L(\hat{w}_m, q)-\hat{L}_m(\hat{w}_m, q))^2+8\max_{q\in\mathcal{Q}}(\hat{L}_m(w, q)-L(w, q))^2
 \\&&+8\max_{q\in\mathcal{Q}}L^2(w, q)+2\max_{w\in\mathcal{W}}\min_{q\in\mathcal{Q}}L^2(w, Q_{\pi_e}-q)
 \\&\leq& 4\max_{w\in\mathcal{W}, q\in\mathcal{Q}}(L(\hat{w}_m, q)-\hat{L}_m(\hat{w}_m, q))^2+8\max_{w\in\mathcal{W},q\in\mathcal{Q}}(\hat{L}_m(w, q)-L(w, q))^2
 \\&&+8\min_{w\in\mathcal{W}}\max_{q\in\mathcal{Q}}L^2(w, q)+2\max_{w\in\mathcal{W}}\min_{q\in\mathcal{Q}}L^2(w, Q_{\pi_e}-q)
 \\&=&12\max_{w\in\mathcal{W}, q\in\mathcal{Q}}(L(\hat{w}_m, q)-\hat{L}_m(\hat{w}_m, q))^2+8\min_{w\in\mathcal{W}}\max_{q\in\mathcal{Q}}L^2(w, q)+2\max_{w\in\mathcal{W}}\min_{q\in\mathcal{Q}}L^2(w, Q_{\pi_e}-q),
 \eeqnn
and therefore
 \beqnn
\E(I_1^2)&\leq& 12\E(\max_{w\in\mathcal{W}, q\in\mathcal{Q}}(L(\hat{w}_m, q)-\hat{L}_m(\hat{w}_m, q))^2)
\\&&+8\min_{w\in\mathcal{W}}\max_{q\in\mathcal{Q}}L^2(w, q)+2\max_{w\in\mathcal{W}}\min_{q\in\mathcal{Q}}L^2(w, Q_{\pi_e}-q).
 \eeqnn
Using this inequality to replace (\ref{Dci1}) and then repeating the discussion in Theorem \ref{Theorem 3}, we can readily get the desired result.\qed

\section{Algorithm Supplement }\label{algorithm supplement}

Algorithm 1 summarizes the pseudo-codes of our MWLA algorithm applied to the taxi environment in Section \ref{section6}.

The algorithm needs the following notation:
\[G=\begin{bmatrix}
0 &\dots &0\\
& \ddots & \vdots\\
0 &&0
\end{bmatrix}_{nh\times nh},\]
$Frequency=[0,\dots,0]_{nh}^{\top},$ \quad
$auxi=[0, \dots, 0]_{nh}^{\top}, \quad \hat{\mu}=[0,\dots, 0]_{nh}^{\top},$
$X$ represents absorbing state set, $Y=\{ h \times i+j, i\in X, j\in \mathcal{A}\}.$

\begin{algorithm}

		\caption{Tabular case}\label{algorithm}
		\begin{algorithmic}[1]
		\Require
		 Off-policy data $D={\{s_{0}^{i},a_{0}^{i},r_{0}^{i},s_{1}^{i},\cdots,s_{T_{i}\wedge H-1}^{i},
        a_{T_{i}\wedge H-1}^{i},r_{T_{i}\wedge H-1}^{i},s_{T_{i}\wedge H}^{i}\}}_{i=1}^{m}$ from the behavior policy $\pi_{b}$;
        a target policy $\pi$ for which we want to estimate the expected return.
        \State{
            {\bf Estimate} the initial state distribution $\hat{\mu}(s)=\frac{1}{m}\sum\limits_{i=1}^{m}1_{\{s_{0}^{i}=s\}}$,
                where $1_{\{\cdot\}}$ is an indicative function.
             \For {episode in D}
                 \For {s,a,$s^{'}$,r in episode}
                 \State $cur=h\times s+a$,
                 \State $G[cur, h\times s^{'}:h\times(s^{'}+1)]+=\pi[s^{'},:],$
                 \State $G[cur,cur]-=1.0,$
                 \State $ Frequency[cur]+=1.$
                 \EndFor
             \EndFor
        {\bf end for}\\
        $auxi=\sum_{s,a}\hat{\mu}(s)\pi(a|s){\bf 1_(s,a)}.$\\
        $tvalid=where(Frequency>0)$  indicates the index of an element whose value is greater than 0.\\
         $\hat{d}_{\pi_{b}}=delete(Frequency,Y,0),$ delete the row corresponding to the absorbing state from Frequency.\\
        $tvalid1=where(\hat{d}_{\pi_{b}}>0).$\\
        $\hat{d}_{\pi_{b}}=\hat{d}_{\pi_{b}}/m.$\\
        ${\hat{\bm G}}=delete(G,Y,0)$, delete the row corresponding to the absorbing state from G.\\
        ${\hat{\bm G}}=delete(\hat{\bm G},Y,1)$, delete the column corresponding to the absorbing state from $\hat{\bm G}$.\\
        ${\hat{\bm G}}[:,tvalid1]={\hat{\bm G}}[:,tvalid1]/(m\times\hat{d}_{\pi_{b}}[tvalid1]).$\\
        $\hat{b}=delete(auxi,Y,0).$\\
        {\bf Compute} $\hat{{\bm u}}
        =\arg\min\limits_{{{\bm u}\geq 0}}\parallel ({\hat{\bm G}+\lambda \bf I})^{\top}{\bm u}+\hat{\bm b}\parallel^{2}$,
        where
        $${\hat{\bm G}}=\frac{1}{m}\sum_{i=1}^{m}\sum_{t=0}^{T_i\wedge H-1}\frac{{\bf 1}_{(s^i_t, a^i_t)}\Big[\sum_{a\in\A}\pi(a|s^i_{t+1}){\bf 1}^{\top}_{(s_{t+1}^i, a)}-{\bf 1}^{\top}_{(s_t^i, a^i_t)}\Big]}{\hat{d}_{\pi_{b}}(s_{t}^{i},a_{t}^{i})}.$$
        $$\hat{\bm b}=\sum_{(s,a)\in\S_0\times\A}\hat{\mu}(s)\pi(a|s){\bf 1}_{(s,a)},$$
        $\lambda$ is a regularization factor, $\bf I$ denotes an identity matrix.\\
        {\bf Parameterize} $w(tvalid)=\frac{{\hat{\bm u}(tvalid1)}}{\hat{d}_{\pi_{b}}(tvalid1)}.$
        \For {episode in D}
            \For {s, a, $s^{'}$, r in episode}
             \State $cur=h\times s+a$,
                 \State $\hat{R}_{\pi_e,m}=\frac{1}{m}\sum\limits_{i=1}^{m}\sum\limits_{t=0}^{T_{i}\wedge H-1}w(cur)\times r.$
                 \EndFor
             \EndFor
        }
        {\bf end for}
        \Ensure $\hat{R}_{\pi_e,m}.$
		\end{algorithmic}	
\end{algorithm}

The idea of the algorithm is explained in Example \ref{ex3.1}. Here, for the convenience of computations,  we set $$w(s,a)=\frac{\mathbf{1}_{\{s,a\}}^{\top}{\bm u}}{\hat{d}_{\pi_{b}}(s,a)}\;\text{and}\;\; q(s,a)=\mathbf{1}_{\{s,a\}}^{\top}{\bm v},\qquad\forall s\in\mathcal{S}_{0},a\in\mathcal{A}.$$
We also introduce a regularization factor $\lambda>0$ which helps us find  the unique solution of the constrained quadratic programming problem $\arg\min\limits_{{{\bm u}\geq 0}}\parallel ({\hat{\bm G}+\lambda \bf I})^{\top}{\bm u}+\hat{\bm b}\parallel^{2}$. When $\lambda$ is sufficient small, the solution is an approximation of $-({\hat{\bm G}}^+)^{\top}{\bm b}$ where ${\hat{\bm G}}^+$ is the Moore-Penrose pseudo-inverse of ${\hat{\bm G}}$. In our experiments, $\lambda$ is set to be $0.001$. 
\end{document}